\documentclass[lettersize,journal]{IEEEtran}
\usepackage{amsmath,amsfonts}
\usepackage{tabularx}
\usepackage[linesnumbered,ruled,vlined]{algorithm2e}
\usepackage{array}
\usepackage[caption=false,font=normalsize,labelfont=sf,textfont=sf]{subfig}
\usepackage{textcomp}
\usepackage{stfloats}
\usepackage{url}
\usepackage{verbatim}
\usepackage{graphicx}
\usepackage{cite}
\usepackage{float}
\usepackage{booktabs}
\usepackage{multirow}
\usepackage{makecell}
\usepackage{pifont}
\newcommand{\xmark}{\ding{55}}
\newcommand{\cmark}{\ding{51}}
\newcolumntype{Y}{>{\centering\arraybackslash}X}
\usepackage{color}
\usepackage{hyperref}
\usepackage{threeparttable}
\begin{document}

\title{Multi-task SAR Image Processing via GAN-based Unsupervised Manipulation}
\author{Xuran Hu,~\IEEEmembership{Student,~IEEE},
	Mingzhe Zhu,
	Ziqiang Xu,
	Zhengpeng Feng,
	and Ljubi\v{s}a Stankovi\'c,~\IEEEmembership{Fellow,~IEEE}

	\thanks{X. Hu, Z. Feng, Z. Xu and M. Zhu are with School of Electronic Engineering, Xidian University, Xi’an, China. Email: XuRanHu@stu.xidian.edu.cn, zpfeng\_1@stu.xidian.edu.cn, zhumz@mail.xidian.edu.cn}
	\thanks{X. Hu and M. Zhu are also with Kunshan Innovation Institute of Xidian University, School of Electronic Engineering, Xidian University, Xi'an , China}
	\thanks{L. Stankovi\'c is with the EE Department of the University of Montenegro, Podgorica, Montenegro. Email: ljubisa@ucg.ac.me.}
	\thanks{Corresponding author: Mingzhe Zhu}}

\markboth{Journal of \LaTeX\ Class Files,~Vol.~14, No.~8, August~2021}%
{Shell \MakeLowercase{\textit{et al.}}: A Sample Article Using IEEEtran.cls for IEEE Journals}

\IEEEpubid{0000--0000/00\$00.00~\copyright~2021 IEEE}

\maketitle

\begin{abstract}
Generative Adversarial Networks (GANs) have shown tremendous potential in synthesizing a large number of realistic SAR images by learning patterns in the data distribution. Some GANs can achieve image editing by introducing latent codes, demonstrating significant promise in SAR image processing. Compared to traditional SAR image processing methods, editing based on GAN latent space control is entirely unsupervised, allowing image processing to be conducted without any labeled data. Additionally, the information extracted from the data is more interpretable. This paper proposes a novel SAR image processing framework called GAN-based Unsupervised Editing (GUE), aiming to address the following two issues: (1) disentangling semantic directions in the GAN latent space and finding meaningful directions; (2) establishing a comprehensive SAR image processing framework while achieving multiple image processing functions. In the implementation of GUE, we decompose the entangled semantic directions in the GAN latent space by training a carefully designed network. Moreover, we can accomplish multiple SAR image processing tasks (including despeckling, localization, auxiliary identification, and rotation editing) in a single training process without any form of supervision. Extensive experiments validate the effectiveness of the proposed method.
\end{abstract}


\section{Introduction}

\IEEEPARstart{S}{ynthetic} aperture radar (SAR) has gained significant popularity in various fields such as remote sensing, electronic reconnaissance, and disaster rescue, owing to its powerful imaging capabilities regardless of day or night and weather conditions. However, SAR imaging is an expensive means of imaging that relies on aerospace surveying and mapping. The cost of obtaining a large number of SAR images is high. In order to meet the requirements of big data and deep learning, many works attempt to generate realistic SAR images using generative networks. Common methods include Variational Autoencoders (VAE) \cite{qian2022learning, jin2020mmfall}, Generative Adversarial Networks (GANs) \cite{goodfellow2014generative, doi2020gan}, and Diffusion Models (DDPM) \cite{ho2020denoising, perera2023sar, hu2024sar}. With the development of generative networks, many researchers have found that controlling image attributes by introducing a set of latent codes can effectively achieve SAR image editing. Representative methods include infoGAN \cite{chen2016infogan} and StyleGAN \cite{karras2020analyzing}, 

GAN-based approach for SAR image processing which achieved through editing latent codes is often unsupervised \cite{shen2021closed} or weakly supervised \cite{shen2020interpreting}. Faced with complex and diverse data, generative networks summarize rules by exploring the intrinsic distribution of data to find meaningful directions. Taking SAR despeckling as an example, most deep learning-based methods require the construction of image pairs of noise and ground truth for supervised learning. This entails a large amount of annotation. Additionally, since noise-free SAR images do not exist in reality \cite{hu2023research}, artificial noise synthesis is often required, affecting the interpretability of the model. In contrast, Latent space-based despeckling methods seek the transformation rules from high noise distribution to low noise distribution in data, thereby achieving more reliable SAR despeckling.

\begin{figure}[!t]
	\centering
	\includegraphics[width=\linewidth]{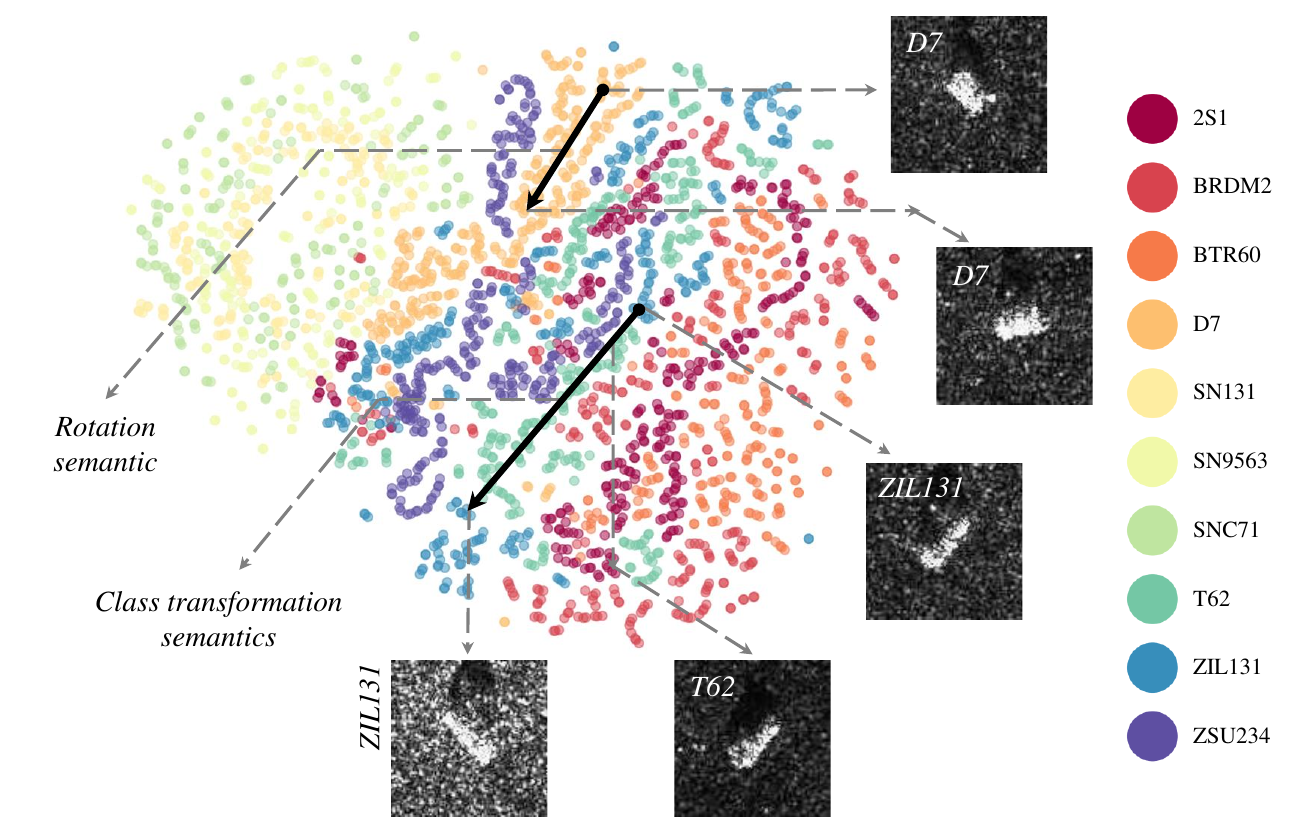}
	\caption{Visualization of t-SNE \cite{van2008visualizing} for latent space samples. GAN latent space contains a wealth of semantic information, and moving along semantic directions in the latent space enables editing.}
	\label{fig_1}
\end{figure}

Another advantage of this technology is that it allows us to build a comprehensive model for SAR image processing, capable of performing multiple tasks. Due to the abundance of semantic information in GANs' latent space, different directions can achieve completely different functions. InterGAN \cite{shen2020interpreting} indicates the existence of hyperplanes in latent space that can separate different categories, and operating on latent vectors along the normal vectors of the hyperplanes can achieve specified feature transformations. As shown in figure \ref{fig_1}, these classifications can include high noise/low noise, target types, target poses, and so on. As long as there are \IEEEpubidadjcol boundaries in high-dimensional data, we can use latent vector manipulation to achieve category transformations (i.e., image despeckling, target category conversion, target pose transformation).

Despite its enormous potential, this technology has not yet been widely used in the field of SAR image processing. Some similar ideas have made progress using self-supervised learning \cite{dalsasso2021if, wang2023unsupervised} and other methods, but they are only designed for specific tasks. The biggest challenge facing GAN-based SAR image processing is how to identify meaningful directions in GANs' latent space. As mentioned above, the latent space of generative networks contains a large number of semantic directions, but these directions are often interrelated. Failure to find appropriate semantic directions can lead to semantic entanglement, seriously affecting the accuracy and robustness of SAR image processing. Therefore, decoupling directions in latent space has become an important topic \cite{zhu2023linkgan, liu2023delving}.

To address the above-mentioned issues, this paper proposes a novel SAR image processing framework based on the latest StyleGAN model, called GAN-based Unsupervised Editing (GUE), aiming to achieve the following two objectives: (1) decoupling directions in latent space through unsupervised learning to enable more effective SAR image editing; (2) establishing prototype of a comprehensive SAR image processing model that can perform multiple tasks with a single training process. Specifically, to identify interpretable directions within the latent space, we train a network to reconstruct orientations perturbed by latent codes. Decoupling the directions ensures that the semantics of interpretable direction operations remain consistent. Subsequently, we select different semantic directions to accomplish various tasks.

Our contributions can be summarized as follows: 

\begin{enumerate} 
	\item GUE represents the first attempt to explore GAN latent space for SAR image processing. We employ a completely unsupervised approach to explore potential semantic directions in the latent space and achieve interpretable SAR image editing.
	\item We designed a direction decoupling network to ensure that the directions in optimization are linearly independent, thereby effectively exploring the semantic directions in GANs' latent space.
	\item We enhanced the interpretability of image editing by implementing transparent semantic operations, resulting in more reliable SAR image processing.
	\item We provide a novel perspective on SAR image processing and attempt to establish a prototype of a comprehensive model. GUE achieves semantic decoupling through a single training session and performs multiple SAR image processing tasks, including but not limited to SAR despeckling, SAR background removal, editing SAR image rotation to enhance image understanding, and guiding SAR ATR.
\end{enumerate}

The remainder of this paper is structured as follows: Section \ref{Section_2} delves into the foundational work, encompassing Generative Adversarial Networks (\ref{Section_2_1_1}) and properties of GANs' latent spaces (\ref{Section_2_1_2}), as well as SAR image processing tasks such as despeckling (\ref{Section_2_2_1}), segmentation (\ref{Section_2_2_2}), and rotation (\ref{Section_2_2_3}). Section \ref{Section_3} provides a comprehensive introduction of GUE implementation. In Section \ref{Section_4}, we present the experimental results across multiple tasks, including SAR despeckling (\ref{Section_4_2}), SAR background segmentation (\ref{Section_4_3}), SAR rotational editing (\ref{Section_4_4}), and guided SAR target recognition (\ref{Section_4_5}). This section also includes ablation studies (\ref{Section_4_6}) to validate the effectiveness of GUE module design. Section \ref{Section_5} discusses the limitations of our approach (\ref{Section_5_1}) and potential avenues for future research (\ref{Section_5_2}). Finally, Section \ref{Section_6} concludes with a summary of our findings.

\section{Related Work}\label{Section_2}

\subsection{Generative Adversarial Networks and Latent Space}

\subsubsection{Generative Adversarial Networks}\label{Section_2_1_1}

\begin{figure*}[!t]
	\centering
	\includegraphics[width=\textwidth]{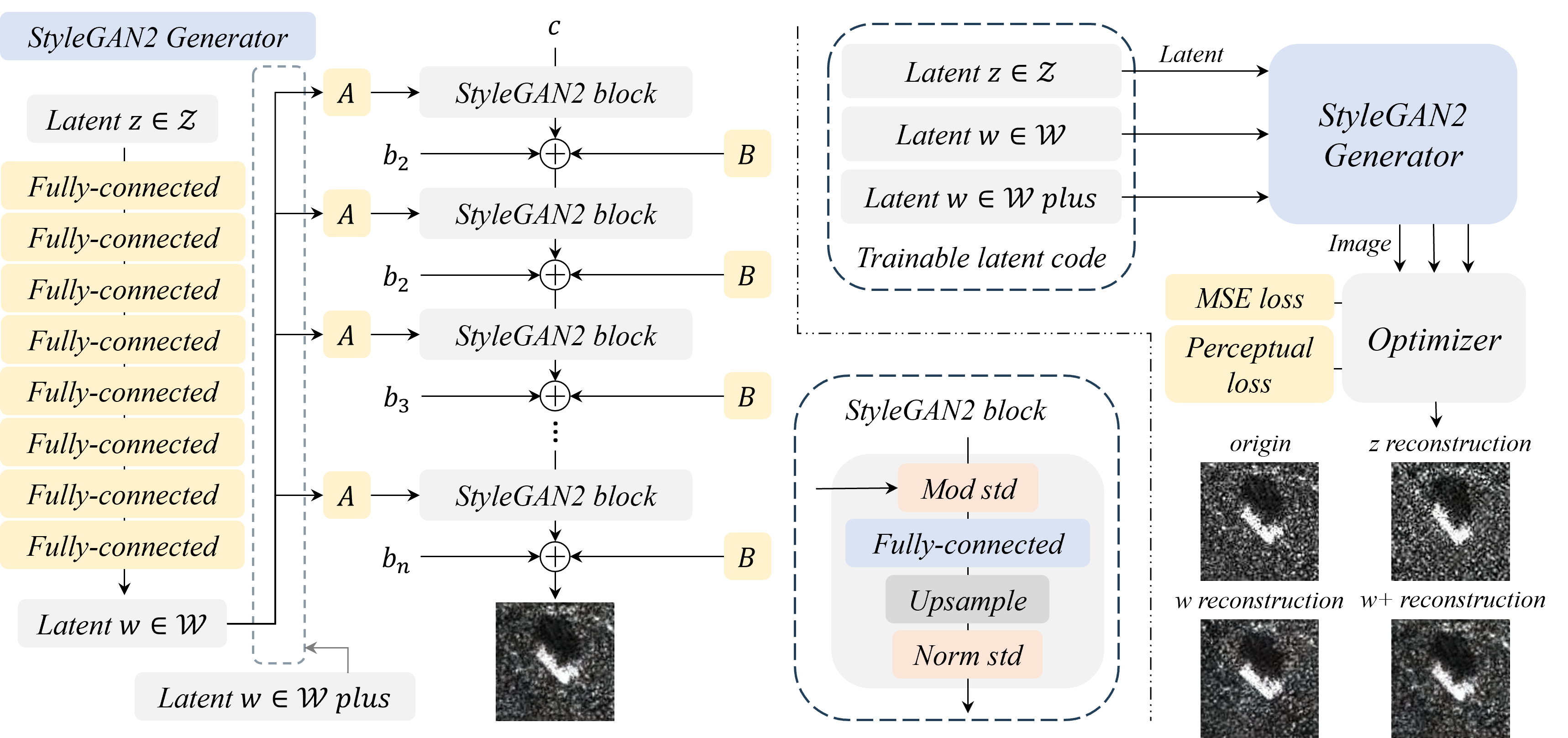}
	\caption{Left: The structure of the StyleGAN2 \cite{karras2020analyzing} generator, containing three different latent spaces: the original input space denoted as $z$, the intermediate latent space represented as $w$, and the modulated latent space $w^{+}$. Right: The Image2StyleGAN \cite{abdal2019image2stylegan, wei2022e2style} algorithm, corresponding to the optimized reconstruction outcomes of the three latent spaces, demonstrates deformations in the $z$ and $w$ space reconstruction results, while the $w^{+}$ space reconstruction most faithfully reverts the original image.}
	\label{fig_2}
\end{figure*}

GAN was proposed by Goodfellow et al., comprising of a generator network, $G$, and a discriminator network, $D$. The generator manages to approximate the real data distribution from a random distribution, and the discriminator estimates the probability that the input sample is a real image or synthesized by generator. Specifically, for an image, $x$ from the training set, the training process is conducted by optimaizing the loss function as follows:

\begin{equation}
	\mathop{\min}\limits_G  \mathop{\max}\limits_D \mathcal L(G, D) =  \mathbb{E}_x\{\log D(x)\}  + \mathbb{E}_z\{\log(1 -  D(G(z)))\}\label{eq:lossgan}
\end{equation} 
where $z$ is a random noise vector. By minimizing (\ref{eq:lossgan}), the discriminator will not be able to discriminate the real image from training set or the synthesized images from the generator, meaning it will
produce the output $D(x) = D(G(z))=\frac{1}{2}$.

However, no restrictions are imposed on noise vector, $z$, in (\ref{eq:lossgan}), thus, it is difficult to further edit the details of synthesized images. To tackle this issue, many advancded GAN models, such as InfoGAN, BigGAN, StyleGAN, etc.\cite{liu2023delving,pehlivan2023styleres,baykal2023clip}, introduce latent codes to learn more detailed properties of generated images.  InfoGAN is a complete unsupervised model, aiming to learn interpretable and disentangled representations by explicitly maximizing the mutual information between latent codes and generated images.  

StyleGAN is a state-of-the-art GAN model that excels in generating high-quality and diverse images. It introduces a novel architecture that separates the control of image content and style, allowing for fine-grained manipulation of generated images. By disentangling the latent space into style and content components, StyleGAN enables users to control various aspects of the generated images, such as facial attributes \cite{jiang2023styleipsb,yang2023high}, background, and artistic style. 

The structure of StyleGAN is depicted in Figure \ref{fig_2} (left). Due to its robust generative capabilities and the introduction of style vectors for image manipulation, our research achieve SAR despeckling and editing by identifying meaningful semantic directions within its latent space.

\subsubsection{Semantics in GAN’s Latent Spaces}\label{Section_2_1_2}

The relation between specific semantics and latent space has been a heated research direction in image synthesis of GANs. Radford et al. \cite{radford2015unsupervised} demonstrates that the generator possesses intriguing vector arithmetic properties, allowing for easy manipulation of numerous semantic qualities of the generated samples. Subsequently,  \cite{goetschalckx2019ganalyze, shen2020interpreting} utilize supervised learning strategy to identify semantics within the latent space. They successfully resolved the issue of direction entanglement by projecting the GAN's latent space into hyperplanes and creatively deploy a network classifier to avoid manual annotation. However, these supervised learning-based methods are limited in extrapolating semantics beyond the given annotations.

To alleviate the drawbacks of supervised methods, some unsupervised methods are proposed to represent the relation between semantic and latent space spontaneously in adversarial training. For example, \cite{shen2021closed} introduced a closed-form decomposition of GAN latent space semantics as well as their  varying directions unsupervisely identified by decomposing pre-trained network weights. \cite{voynov2020unsupervised, tzelepis2021warpedganspace} developed reconstructors capable of restoring the displacement transformation of the latent code. By appropriate optimization, they achieve the decomposition of sematcis in latent space.

GANs can effectively learn data distributions and, compared to other generative models like VAEs \cite{kingma2013auto}, can achieve high-quality reconstructions. Some generative models, such as DDPMs, can also perform image editing through latent space manipulation \cite{shih2021ganmex, zhang2022unsupervised, park2023understanding, kwondiffusion}. However, this often requires restructuring the network, and some studies have shown that introducing additional latent codes can degrade image generation quality. In this study, we use StyleGAN \cite{tzelepis2021warpedganspace} as the backbone network to achieve SAR editing by manipulating the reconstructed latent space. Building on previous concepts, we orthogonalized latent codes to maximally achieve semantic disentanglement. What sets our approach apart from prior works is the utilization of a novel transformer for orthogonalizing latent codes, and extending the model into the $w^+$ space for more effective editing of real images. Applying this method to SAR despeckling and editing presented a novel direction, given the absence of real ground truth noise-free images for training, introducing a new avenue for unsupervised despeckling in SAR images.

\subsubsection{GAN Inversion}\label{Section_2_1_3}

Despite the success of GANs in image synthesis, applying pretrained GAN models to real image processing remains challenging\cite{wang2022high,roich2022pivotal}. The usual approach is to invert the given image back to latent codes so that the generator can reconstruct it. Existing methods for inverting the generation process can be categorized into two types. One type directly optimizes the latent codes by minimizing the reconstruction error through forward propagation \cite{richardson2021encoding}. The other type involves training an additional encoder to learn the mapping from the image space to the latent space. However, training-based methods yield unsatisfactory reconstruction results, especially when dealing with high-resolution images.

\begin{figure*}[!t]
	\centering
	\includegraphics[width=\textwidth]{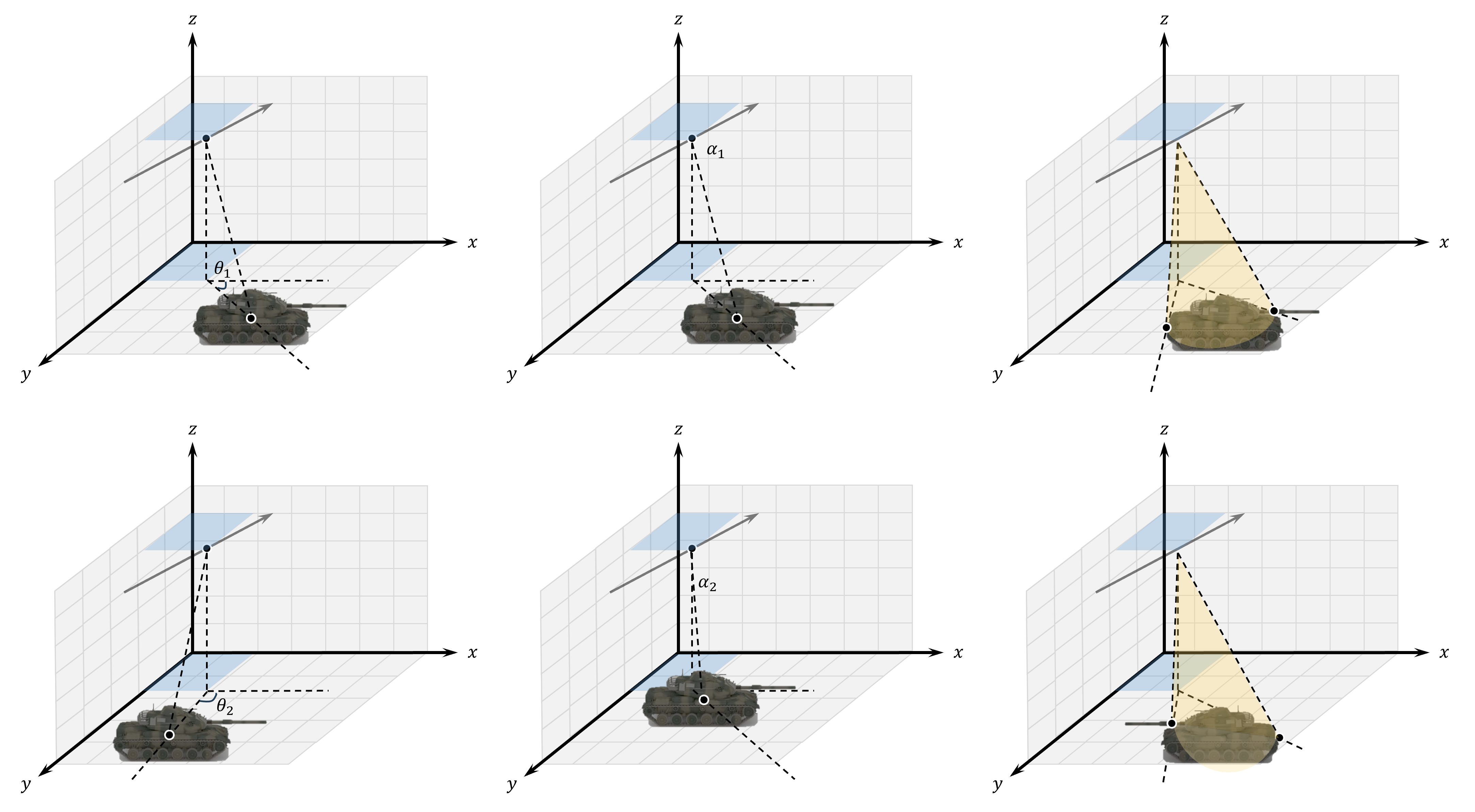}
	\caption{Compared with optical images, SAR images are more sensitive to rotation transformation. Standard rotation semantics include (1) azimuth rotation (first column); (2) pitch angle rotation (second column); (3) attitude transformation (third column).}
	\label{fig_3}
\end{figure*}

In this study, we employ Image2StyleGAN \cite{abdal2019image2stylegan, wei2022e2style} to map the image into the latent space and obtain the corresponding latent code for image editing.
Image2StyleGAN is an optimization-based GAN inversion algorithm that employs an optimized model to reconstruct the original image. The structure of Image2StyleGAN is depicted in Figure \ref{fig_2} (right). The traditional Mean Squared Error (MSE) loss is introduced in the optimization model. However, pixel-level MSE loss alone cannot achieve high-quality embedding. Therefore, the perceptual loss \cite{johnson2016perceptual} is introduced into the optimization model as a regularization term. The loss function of Image2StyleGAN is defined as follows:

\begin{equation}
	z^*=\min _{\mathrm{z}} L_{percept}(G(z), I)+\frac{1}{N}\|G(z)-I\|_2^2,
\end{equation} among the equations, $I$ represents the input image, $G(\cdot)$ denotes the pretrained generator, and $N$ represents the dimension of the input image. $L_{percept}(\cdot,\cdot)$ represents the perceptual loss, and its specific expression is given by:

\begin{equation}
	L_{percept}\left(I_1, I_2\right)=\sum_{j=1}^K \frac{1}{N_j}\left\|F_j\left(I_1\right)-F_j\left(I_2\right)\right\|_2^2
\end{equation}

In the equation, $F_j(I)$ represents the feature map of the specified convolutional layer in VGG16 network, with $I$ as the input, and $N_j$ represents the dimension of the feature map.

In the synthetic image experiments of this paper, we conducted unsupervised despeckling and editing based on the $z$-space by training and testing using randomly generated latent codes. For real SAR images, to manipulate them in the latent space, we first employed Image2StyleGAN to map images back to the latent space.

\subsection{SAR Image Processing and SAR ATR}

\subsubsection{SAR Despeckling}\label{Section_2_2_1}

SAR images often suffer from speckle noise, which arises from the coherent superposition of radar echoes. These noise severely degrades the quality of SAR images and poses challenges for their practical utilization \cite{lee1999polarimetric}. Consequently, despeckling techniques have become increasingly heated in SAR image processing. 

Deep-learning-based methods recently obtained great success in SAR despeckling. SAR-CNN \cite{chierchia2017sar, tucker2022polarimetric} employs residual learning, which significantly enhances convergence speed. ID-CNN \cite{wang2017sar} and SAR-DRN \cite{zhang2018learning} emphasize the optimization of model depth and size to obtain better performance. \cite{vitale2020multi} and InSAR-MONet \cite{vitale2022insar} proposed new loss function to analysis image's statistical properties. SAR2SAR \cite{dalsasso2021sar2sar}, SAR-IDDP \cite{lin2022self}, and MRDDANet \cite{liu2022mrddanet}, integrate considerations pertaining to the spatial correlation intrinsic to SAR data.

These supervised approaches, however, require a large number of precisely annotated SAR images with different levels of speckle noise for supervised learning. These methods require a substantial amount of annotation for training. However, ground truth noise-free images do not exist in reality. This limitation hinders these methods from adaptively handling different levels and types of noise. Therefore, some works like MERLIN \cite{dalsasso2021if, wang2023unsupervised} have introduced self-supervised or unsupervised learning to address these issues and have made some progress.

Recently, some researches related to generative adversarial networks become popular. Various GANs, such as ID-GAN \cite{wang2017generative}, SAR-GAN \cite{wang2018generating, zeyadaresolving, tan2021serial}, cycle-GAN \cite{zhang2023blind, kumar2023novel, zeyada2022resolving} have been used in SAR despeckling by mining the intrinsic characteristics of speckle noise automatically. Additionally, to provide a more nuanced control over the properties of SAR images, some advanced GANs, such as InfoGAN \cite{chen2016infogan}, BigGAN \cite{brock2018large}, StyleGAN \cite{karras2019style}, manage to manipulate the speckles in the generated SAR images by introducing latent codes. However, the relation between the properties (e.g., the intensity of speckle noise) and latent codes is usually unclear, thus it is hard to generate SAR images with the precise speckle noise level by maneuvering the latent codes.

\subsubsection{SAR Background Segmentation}\label{Section_2_2_2}

SAR background segmentation refers to the separation of targets from background information, which can shield targets from the interference of cluttered background noise. It is an important preprocessing step for SAR target recognition \cite{wang2020deep, gao2024can, peng2023learning}, as extensively demonstrated by numerous studies. Additionally, this technique is crucial for SAR recognition network interpretation \cite{li2023discovering, feng2023analytical, heiligers2018importance, zhu2024unveiling, hasanpour2022unboxing, li2023discovering, cui2024deep}, as segmenting target and background regions and studying their attribution to the network can provide a better understanding of network's decision mechanism.

Most traditional methods achieve background segmentation by adjusting thresholds. Common approaches include histogram-based methods \cite{chandrasekar2020highly}, which extract target regions through pixel histograms, and clustering-based methods \cite{zhang2008spectral}, which segment background and target regions by clustering. These traditional methods rely on parameter selection and cannot adaptively segment SAR images with different noise levels. The introduction of deep learning can significantly reduce parameter dependence, and many deep learning-based methods have been proposed \cite{zhao2023multitask, yasir2023instance}. However, these methods often require manual annotation of SAR target regions for supervised learning, which incurs high costs. Nowadays, more and more research aims to achieve this task through unsupervised or self-supervised methods \cite{feng2021target, presles2023synthetic}, with some efforts being made in this field.

\begin{figure*}[!t]
	\centering
	\includegraphics[width=18cm]{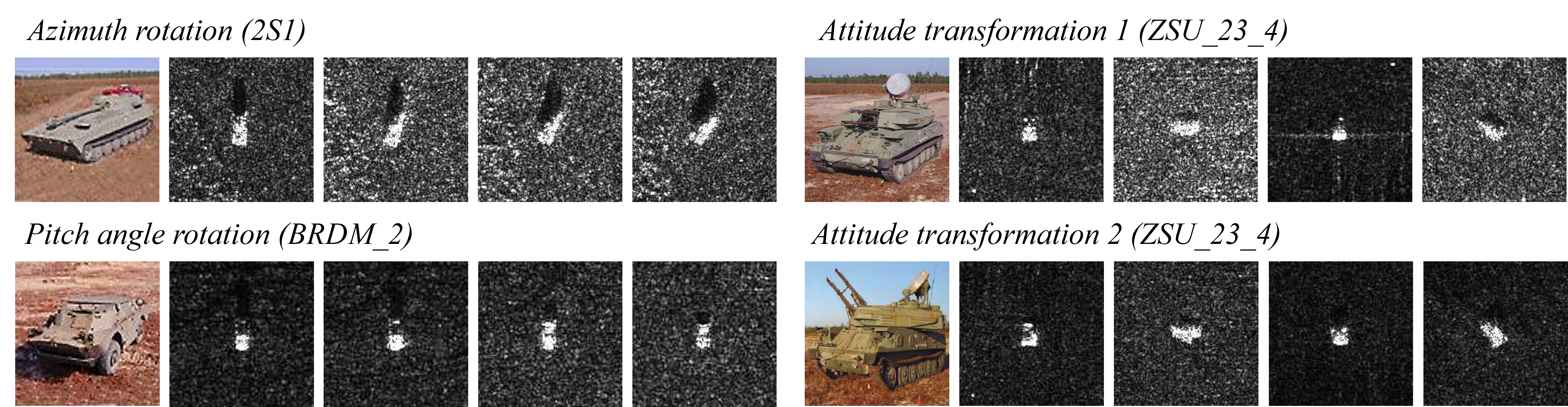}
	\caption{Illustrative Instances of SAR Rotation: These instances encompass the azimuth rotation of the 2S1, the pitch rotation of the BRDM-2, and the attitude transformation of the ZSU-23-4 (which encompasses imaging results depicting two distinct attitudes characterized by varying azimuth angles).}
	\label{fig_4}
\end{figure*}

\subsubsection{SAR Rotation and Data Enhancement}\label{Section_2_2_3}

Recent research has indicated \cite{dong2021keypoint} that SAR images exhibit higher sensitivity than optical images. Minor variations in azimuth, pitch angle, and target orientation can significantly impact SAR imaging due to the distinct characteristics of side-view coherent imaging. The substantial differences in SAR image appearance for the same target arise from the interaction between electromagnetic waves, target structures, and the correlation of echoes originating from different target positions.

Figure \ref{fig_3} illustrates various rotational transformations during SAR imaging: azimuthal (first column), pitch (second column), and attitude transformations (third column). These encompass the target's intrinsic rotation and deformations, such as a tank turret's rotation. Furthermore, Figure \ref{fig_4} showcases changes in SAR images of three distinct ground targets as their azimuth, pitch angles, and attitudes undergo individual transformations.

With the advent of generative adversarial networks (GANs), synthetic SAR image generation has become feasible. However, returning to the initial SAR recognition task, current synthesis methods may encounter two issues: (1) Some methods fail to retain essential SAR image features, such as background and shadow regions, which play a crucial role in SAR target recognition. Losing this information during data augmentation may lead to a decrease in model performance. (2) Some methods allow the synthesis of a large number of SAR images but struggle to extract relevant information from the synthesized data, such as a series of samples depicting SAR rotated along characteristic angles. This makes it challenging to employ such generative models for guiding subsequent SAR recognition tasks and limits their utility to data augmentation.

In response to these challenges, several approaches have been proposed \cite{feng2023interpretation, hu2024manifold}. AGGAN \cite{sun2023attribute} emphasizes inter-object correlations through few-shot generation, while ARGN \cite{sun2019sar} employs attribute-guided transfer learning for data augmentation. In GUE, we adopt a different strategy where we forego the selection of associated samples and instead train the generator directly on the entire dataset. Subsequently, we employ unsupervised learning to identify correlated features that can guide SAR target recognition.

\section{Methodology}\label{Section_3}
\label{section:B}

\begin{figure*}[!t]
	\centering
	\includegraphics[width=18cm]{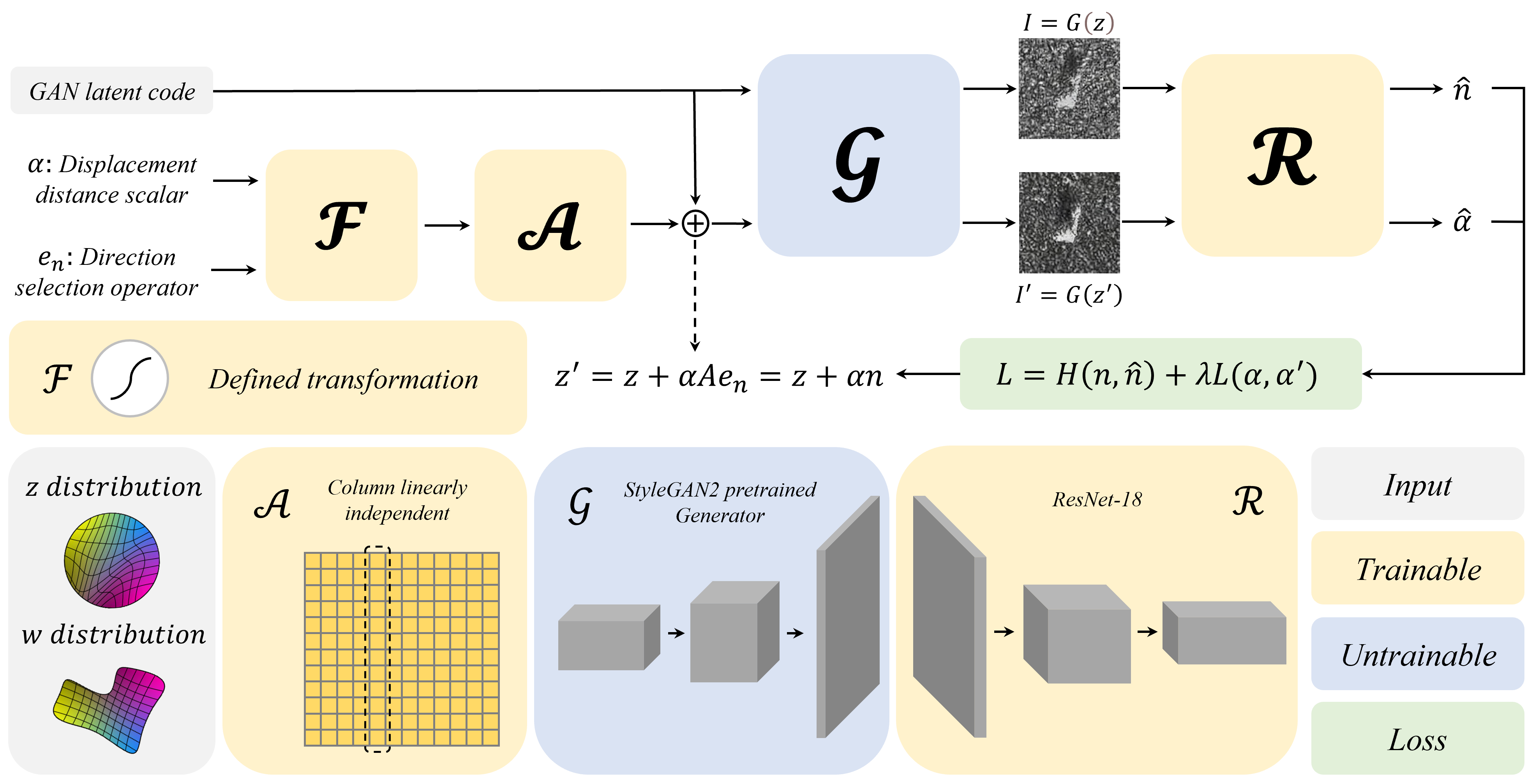}
	\caption{Method overview: Given a pretrained generator $G$, find possible interpretable directions in the GANs$\prime$ latent space. Given a set of latent vectors satisfying a specific distribution \cite{karras2019style}, the displacement operator $z+\alpha A e_n$ is obtained by the direction selection operator and displacement distance operator after defined transformation. The image pair obtained after inputting the original hidden vector $z$ and displacement vector $z^{\prime}$ into $G$ restores the direction index $n$ and displacement distance $\alpha$ through the reconstructor $R$. During the optimization process, $A$ and $G$ are optimized simultaneously, and each column of $A$ is automatically decoupled.}
	\label{fig_5}	
\end{figure*}

This section provides a detailed introduction to GUE. The method's flowchart is presented in Figure \ref{fig_5}. The generator $G$ in GAN framework can be viewed as a mapping from input latent code $z$ to output image $I$, which can be represented as:

\begin{equation}
	I = G(z). 
\end{equation}

We aim to discover an interpretable direction $n$ that allows $z$ to move in that direction, resulting in a modified input:

\begin{equation}
	z^{\prime}=z+\alpha n.
\end{equation}

The interpretability aspect refers to the notion that the transformed output $I^{\prime}=G\left(z^{\prime}\right)$ can be comprehended with the original production $I$.

To begin, we introduce the matrix $A \in \mathbb{R}^{k \times N}$, where $k$ denotes the dimension of GANs' latent space, and $N$ represents the desired number of directions to be identified. The primary objective of GUE is to discover the top $N$ directions corresponding to the columns of matrix $A$, which elicit substantial changes in the image. Simultaneously, to ensure the decoupling of these directions, enabling image editing to be influenced by individual factors only, we aim to enforce orthogonality among each semantic direction in matrix $A$. This orthogonality property facilitates latent code to traverse diverse directions with easily interpretable semantics. Consequently, we impose the condition that matrix $A$ remains orthogonal throughout the optimization process.

Next, we introduce a reconstructor, denoted as $R$, which takes a set of images as input. This set comprises the original image $I=G(z)$ and the image $I^{\prime}=G\left(z^{\prime}\right)$ generated by displacing the latent code $z$ in a specific direction. The role of the reconstructor is to identify the direction $n$ and displacement distance $\alpha$ such that $z^{\prime}=z+\alpha n$. We utilize each column of the matrix $A$ as a direction and apply a displacement operation on $z$ to obtain an image pair:

\begin{equation}
	(I, I^{\prime}) = \left(G(z), G\left(z+\alpha A e_n\right)\right). 
\end{equation}

Here, $e_n$ denotes an identity matrix that preserves only the $n$-th row, and $\alpha$ is a scalar. Specifically, $z+\alpha A e_n$ indicates that $z$ moves a certain distance along the $n$-th column of $A$. The reconstructor aims to discover the displacement transformation within matrix $A$ and reconstruct the transformation between the image pairs. Mathematically, the reconstructor $R$ can be represented by the following mapping: 

\begin{equation}
	R\left(I, I^{\prime}\right)=(n^{\prime}, \alpha^{\prime}).
\end{equation}

To facilitate the reconstructor in accurately restoring the transformation of the latent code, our optimization objective entails defining the following loss function:

\begin{equation}
	\min _{A, R} \mathrm{E}_{z} [H\left(n, n^{\prime}\right)+\lambda L\left(\alpha, \alpha^{\prime}\right)],
\end{equation} Among them, the function $H(\cdot, \cdot)$ represents the cross-entropy function:

\begin{equation}
	H\left(n, n^{\prime}\right)=-\sum_{i=1}^m n\left(x_i\right) \log \left(n^{\prime}\left(x_i\right)\right),
\end{equation} which is utilized to quantify the dissimilarity between the true direction, denoted as $n$, and the predicted direction indicated as $n^{\prime}$. The function $L(\cdot, \cdot)$ represents the least squares loss:

\begin{equation}
	L\left(\alpha, \alpha^{\prime}\right)=\sum_{i=1}^m\left(\alpha_i-\alpha_i^{\prime}\right)^2,
\end{equation} which is utilized to quantify the dissimilarity between the true distance, denoted as $\alpha$, and the predicted direction indicated as ${\alpha}^{\prime}$.

In GUE, matrix $A$ and reconstructor $R$ are jointly optimized. To facilitate the reconstructor $R$ in accurately reconstructing the columns of $A$, columns decoupling is automatically performed during optimization. This ensures that each column corresponds to a single interpretable element, enhancing semantic clarity. The decoupling of semantics is implied within the optimization process. Empirical evidence has demonstrated that GUE consistently identifies the relevant meaningful semantic directions. Generator $G$ remains unchanged throughout the optimization process, reducing the resource consumption associated with retraining.

In accordance with the resolution of SAR dataset, different networks are selected to construct the reconstructor. For MSTAR, the reconstructor $R$ employs ResNet18 structure while expanding the network's first layer to accommodate the image pair's input with six channels. For datasets with higher resolutions, we can adopt more complex network structures. Regarding data initialization, we set the direction $n$ to follow a uniform distribution $U(1, N)$, and we assign a minimum value of 0.5 to the scalar $\alpha$ to prevent the issue of gradient disappearance caused by excessively small $\alpha$ during optimization process. For the loss function, $\lambda$ is set to 0.5.

\textbf{Selection of $F$}: In the experiment, we randomly sample 512 latent codes to form a matrix $M$. To ensure direction diversity and randomness, we employ several different transformers, which are demonstrated in the figure \ref{fig_6}:

\begin{figure}[!t]
	\centering
	\includegraphics[width=\linewidth]{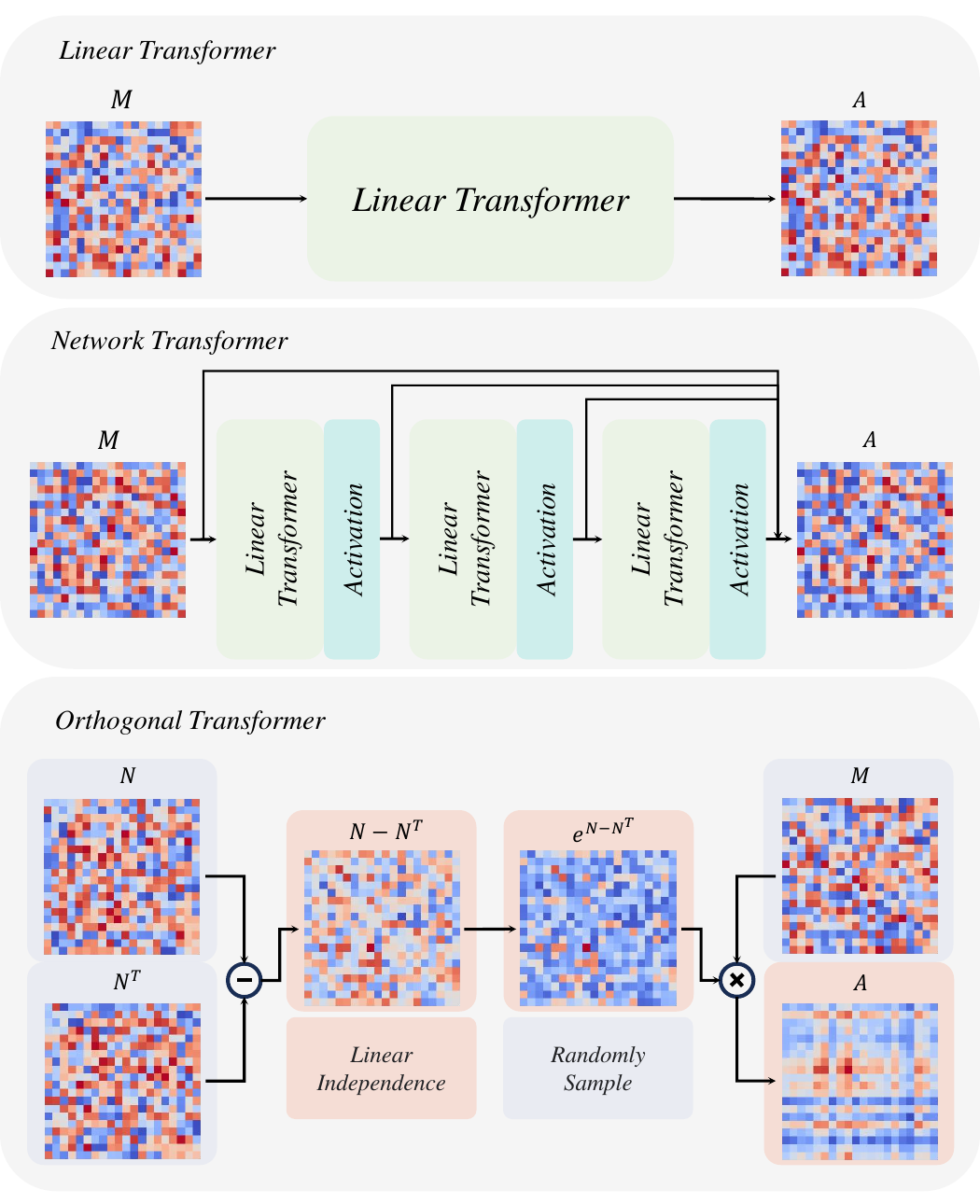}
	\caption{Selection of $F$: three types of transformer structure. From top to bottom: Linear transformer, Network transformer and Orthogonal transformer.}
	\label{fig_6}
\end{figure}

\begin{enumerate} 
	\item Linear Transformer: We define a linear layer to linearly transform the matrix $M$.
	\item Network Transformer: This transformer consists of linear transformation layers, activation layers, and normalization layers, forming a simple mapping network. We input matrix $M$ into the network to obtain its nonlinear mapping result.
	\item Orthogonal Transformer: Unlike linear transformers and network transformers, the direction matrix $A$ processed by $M$ does not satisfy column linear independence. To address this, we define a trainable orthogonal transformer. First, we randomly initialize the matrix $N \in \mathbb{R}^{512 \times 512}$. Then, we define the orthogonal transformation as:
	
	\begin{equation}
		A=\exp \left[M \cdot\left(N-N^T\right)\right].
	\end{equation}
	
	This output matrix satisfies column linear independence, allowing the semantics of each direction to be better decoupled during training.
\end{enumerate}


\textbf{Selection of $A$}: For orthogonal transformer, matrix $A$ must satisfy column linear independence to ensure that each direction controls a single semantic to the greatest extent possible during the optimization process. The number of columns in matrix $A$ represents the desired number $N$ of directions to be identified, while the number of rows corresponds to the dimension of GANs' latent space. By default, we set $N$ to equal the dimension of the latent space, which is 512 for StyleGAN. The StyleGAN latent space encompasses the following three scenarios:

\begin{enumerate} 
	\item $z$ vector: The latent code serves as the original input of StyleGAN. It is denoted as $z \in \mathbb{R}^{1 \times 512}$. 
	\item $w$ vector: The intermediate latent code that transforms $z$ into a new latent space using MLP. It adjusts the data distribution pattern and enhances controllability over the generated image. It is represented as $w \in \mathbb{R}^{1 \times 512}$.
	\item $w^{+}$ vector: An expanded latent code that modulates the $w$ code, enabling each StyleGAN block to correspond to a distinct $w$ input. In typical tasks, the $w^{+}$ vector is a straightforward copy of the $w$ vector. However, in real image editing scenarios, the real image is first reconstructed in the $w^{+}$ space and subsequently edited. In our experiments, ${w^{+}} \in \mathbb{R}^{12 \times 512}$.
\end{enumerate}

Due to the distinct data distributions in three latent spaces of StyleGAN, we conduct separate training for GUE to identify the most prominent orientation under different conditions. For the $z$ space, we utilize the normal distribution $n(1, N)$ as the input for the GAN generator. Regarding the $w$ space and $w^{+}$ space, we extract the latent code corresponding to the reconstructed image and apply kernel density estimation to approximate their respective data distributions. Subsequently, a random code is generated according to the fitted distribution and utilized as input for the GAN generator. The kernel density estimation formula employed is as follows:

\begin{equation}
	\widehat{f}_h(w)=\frac{1}{N} \sum_{i=1}^N K_s\left(w-w_i\right)=\frac{1}{N s} \sum_{i=1}^N K\left(\frac{w-w_i}{s}\right),
\end{equation} among them, $N$ represents the number of samples, $s$ denotes the smoothing parameter responsible for controlling the distribution's bandwidth, and $K(\cdot)$ signifies the kernel function employed.

\begin{algorithm}[!t]
	\SetAlgoNoLine
	\SetKwInOut{Input}{Input}
	\SetKwInOut{Output}{Output}
	
	\Input{
		SAR dataset $\left\{I_k \in \mathbb{R}^{W \times H}, k=1,2 \ldots, N\right\}$\;
		the regularization coefficients $\gamma$\;
	}
	
	\Output{
		Despeckling sample $I^{GUE} \in \mathbb{R}^{W \times H}$\;
	}
	
	train StyleGAN2 on the dataset\;
	build reconstruction network $R$\;
	specify $F$ mode (Linear, Network, Orthogonal)\;
	randomly sample $z$\;

	initialization: network params, matrix $A$, $\alpha$\;

	\For{i $\in$ Iter}{		
		\For{n $\in$ $A\left[c_1, c_2, \ldots, c_N\right]$}{
			$z^{\prime}=z+\alpha n$\;
			$I=G(z)$, $I^{\prime}=G\left(z^{\prime}\right)$\;
			$R\left(I, I^{\prime}\right)=(n^{\prime}, \alpha^{\prime})$\;
			$Loss=\mathrm{E}_{z} [H\left(n, n^{\prime}\right)+\lambda L\left(\alpha, \alpha^{\prime}\right)]$\;
			grad = Adam(loss, params)\;
			optimizer.step()\;
		}
	}
	
	$I^{G U D}=G\left(z+\alpha A\left[c_i\right]\right)$\;
	\caption{GUE (Synthetic sample)}
	\label{algorithm1}
\end{algorithm}

\begin{algorithm}[!t]
	\SetAlgoNoLine
	\SetKwInOut{Input}{Input}
	\SetKwInOut{Output}{Output}
	
	\Input{hyperparameters $s$\;
		the regularization coefficients $\gamma$\;	
		SAR dataset $D = \left\{I_k \in \mathbb{R}^{W \times H}, k=1,2 \ldots, N\right\}$\;
	}
	
	\Output{
		Despeckling sample $I^{GUE}_k \in \mathbb{R}^{W \times H}$\;
	}
	
	train StyleGAN2 on the dataset\;
	build reconstruction network $R$\;
	specify $F$ mode (Linear, Network, Orthogonal)\;
	randomly generate a set of code $z$ and map them to $w$\;
	
	initialization: matrix $A$: \\
	\For{c $\in$ C}{
		$A[:, c]=\frac{1}{N s} \sum_{i=1}^N K\left(\frac{w-w_i}{s}\right)$
	}
	
	initialization: network params, $\alpha$\;
	randomly sample $w$\;

	\For{i $\in$ Iter}{		
		\For{n $\in$ $A\left[c_1, c_2, \ldots, c_N\right]$}{
			$w^{\prime}=w+\alpha n$\;
			$I=G(w)$, $I^{\prime}=G\left(w^{\prime}\right)$\;
			$R\left(I, I^{\prime}\right)=(n^{\prime}, \alpha^{\prime})$\;
			$Loss=\mathrm{E}_{w} [H\left(n, n^{\prime}\right)+\lambda L\left(\alpha, \alpha^{\prime}\right)]$\;
			grad = Adam(loss, params)\;
			optimizer.step()\;
		}
	}
	
	Image2StyleGAN: $I_k \rightarrow w^+_k$\;
	
	$I^{G U D}_k=G\left(w^+_k+\alpha A\left[c_i\right]\right)$\;
	\caption{GUE (Real sample)}
	\label{algorithm2}
\end{algorithm}

\section{Experimental Validation}\label{Section_4}

\begin{figure}[!t]
	\centering
	\includegraphics[width=\linewidth]{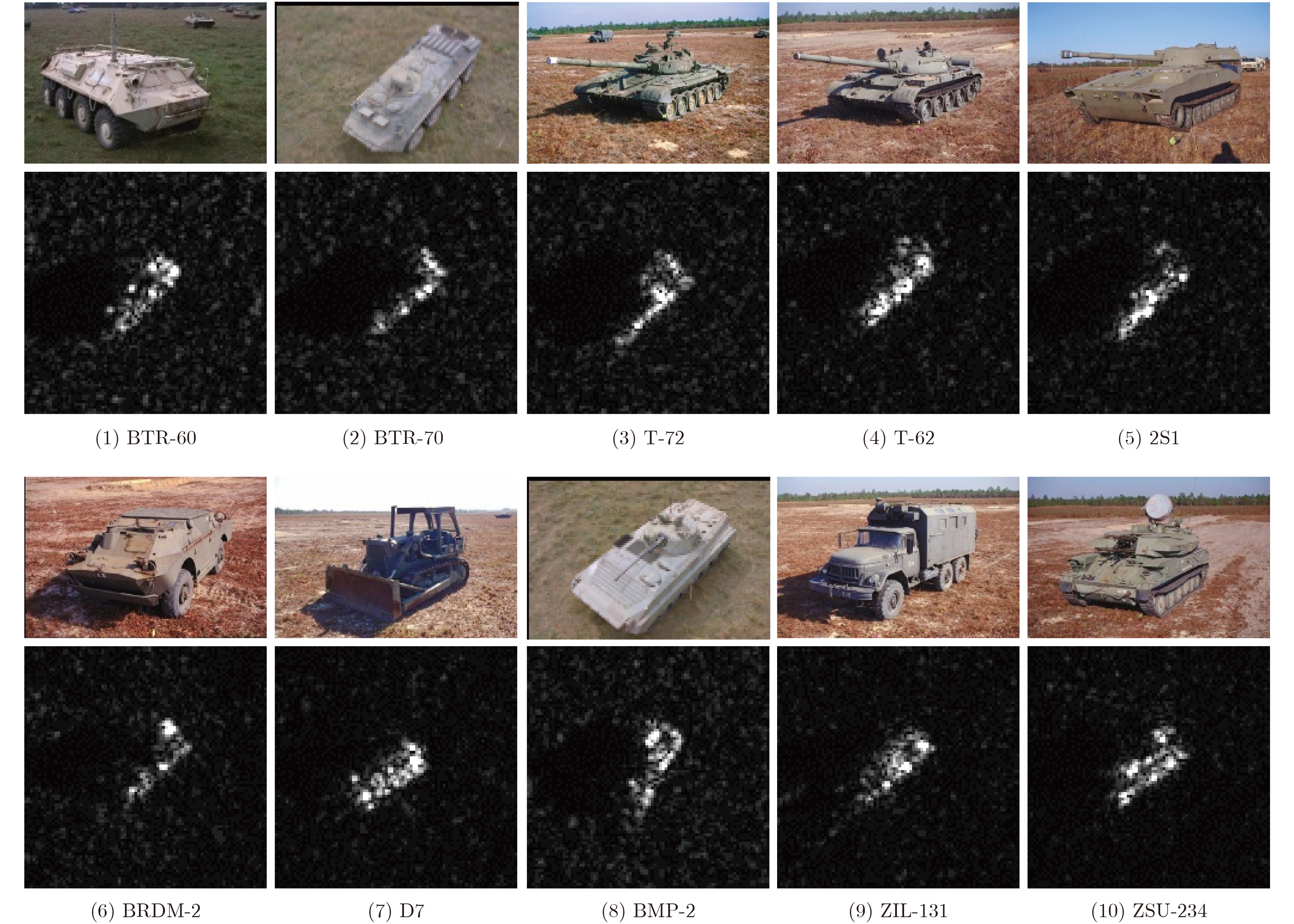}
	\caption{Samples taken from MSTAR datasets.}
	\label{fig_17}
\end{figure}

\begin{figure}[!t]
	\centering
	\includegraphics[width=\linewidth]{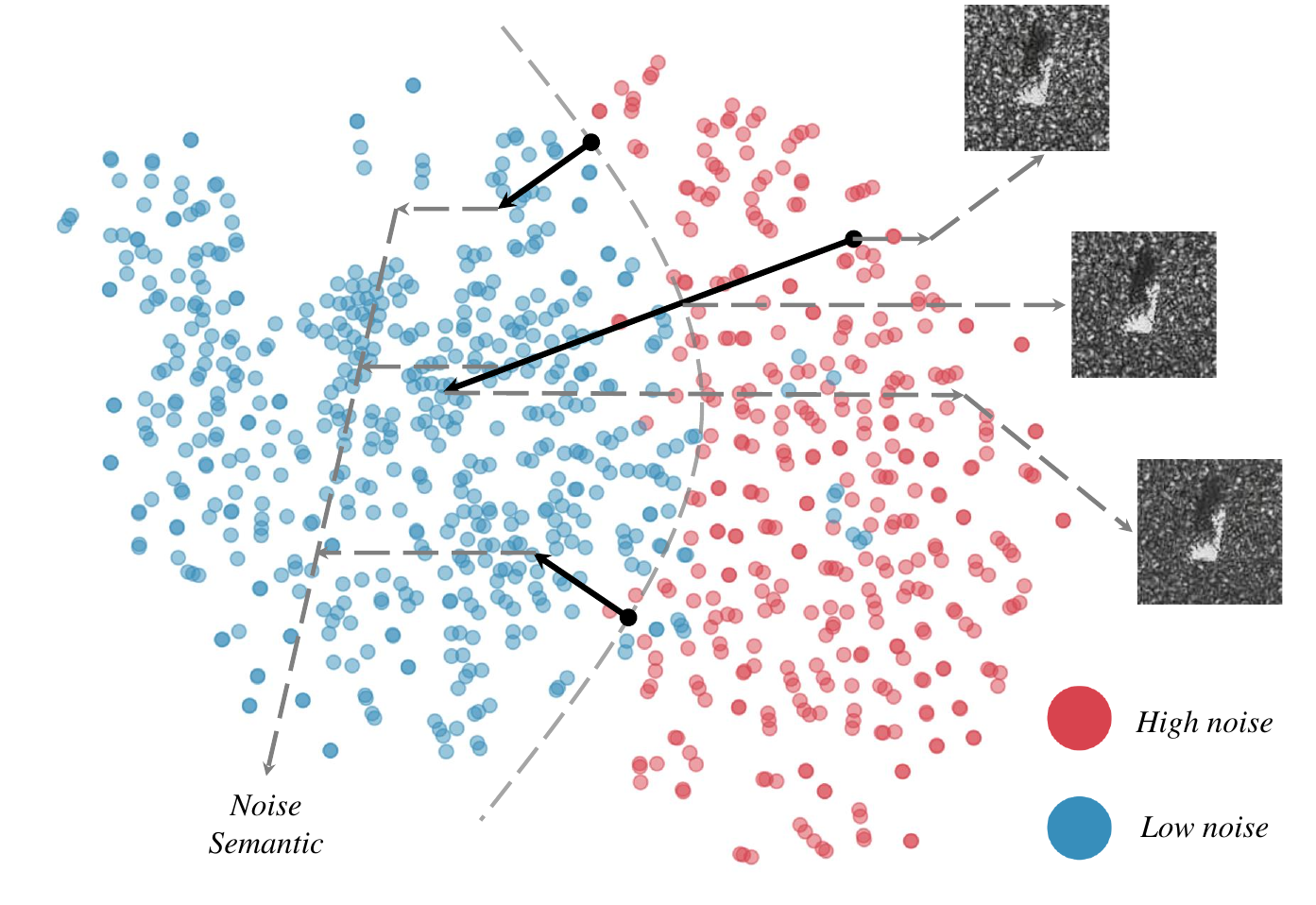}
	\caption{Visualization of t-SNE \cite{van2008visualizing} for latent space samples. GAN latent space contains a wealth of semantic information, and moving along semantic directions in the latent space enables editing.}
	\label{fig_7}
\end{figure}

\begin{figure*}[!t]
		\centering
		\includegraphics[width=0.8\textwidth]{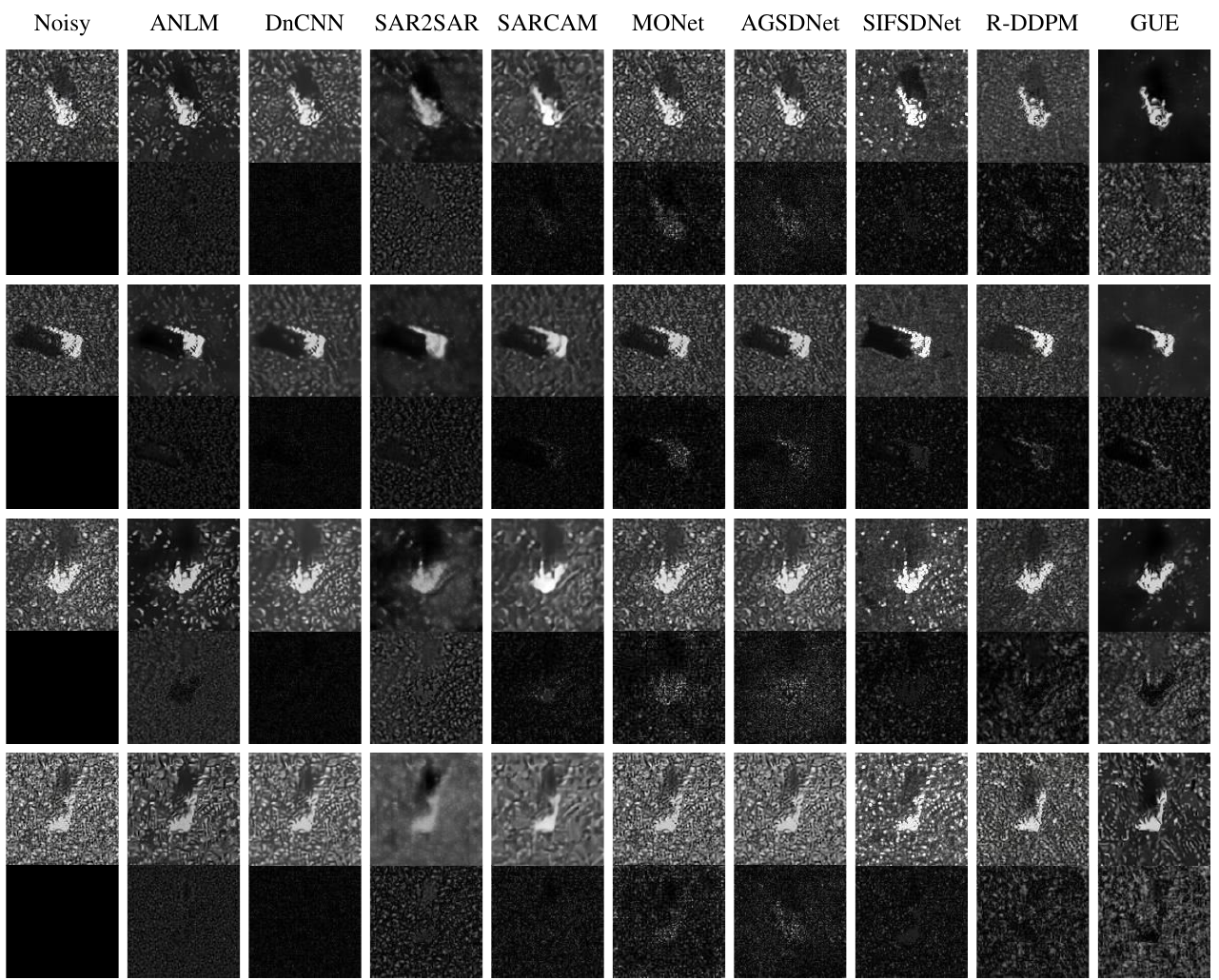}
		\caption{Synthetic image despeckling experiment: left to right: noisy image, ANLM, DnCNN, SAR2SAR, SARCAM, MONet, AGSDNet, SIFSDNet, SARDDPM, GUE, GUE$+$filter. The original noisy image and each method contains three parts, up to down: despeckling image and residual map.}
		\label{fig_8}
	
\end{figure*}

\begin{table*}[!t]
		\caption{Parameter comparison on synthetic image(BEST INDEX VALUES ARE HIGHLIGHTED IN BOLD)}
		\centering
		\newcolumntype{C}{>{\centering\arraybackslash}X}
		\label{tb1}
		\begin{tabularx}{\textwidth}{CCCCCCCC}
			\toprule
			& Metric & ANLM \cite{xiao2020asymptotic} & DnCNN \cite{zhang2017beyond} & SAR2SAR \cite{dalsasso2021sar2sar} & MONet \cite{vitale2021analysis} & InterGAN \cite{shen2020interpreting} & CFF \cite{shen2021closed} \\
			\midrule
			\multirow{3}{*}{(1)} & PSNR & 12.6183 & 10.2462 & 12.4074 & 9.6373 & 6.4239 & 10.7989 \\
			& ENL & 2.4887 & 3.0471 & 4.1884 & \textbf{6.2705} & 1.9986 & 2.3832 \\
			& SSIM & 0.0762 & 0.0748 & 0.0613 & 0.0523 & 0.0359 & 0.0410 \\
			
			\midrule
			
			\multirow{3}{*}{(2)} & PSNR & 12.7700 & 9.9121 & 13.6482 & 8.4130 & 6.4402 & 10.7740 \\
			& ENL & 2.5698 & 2.7700 & 4.2585 & \textbf{8.4932} & 2.2148 & 3.3995 \\
			& SSIM & 0.0815 & 0.0822 & 0.0656 & 0.0418 & 0.0358 & 0.0415 \\
			
			\midrule
			
			\multirow{3}{*}{(3)} & PSNR & 11.7142 & 9.1589 & 11.5931 & 9.1746 & 6.4652 & 10.8264 \\
			& ENL & 2.3639 & 2.6214 & 3.9132 & \textbf{6.6477} & 2.2356 & 3.4159 \\
			& SSIM & 0.0743 & 0.0683 & 0.0579 & 0.0403 & 0.0360 & 0.0426 \\

			\midrule
			
			\multirow{3}{*}{(4)} & PSNR & 12.8423 & 10.3721 & 12.3489 & 9.8511 & 6.5479 & 10.8695 \\
			& ENL & 2.5174 & 2.7928 & 4.2097 & \textbf{8.6403} & 2.1926 & 2.4847 \\
			& SSIM & 0.0735 & 0.0752 & 0.0638 & 0.0407 & 0.0359 & 0.0414 \\
			
			\bottomrule
		\end{tabularx}
		
		\vspace{0.5cm}
		
		\begin{tabularx}{\textwidth}{CCCCCCCC}
			\toprule
			& Metric & SARCAM \cite{9633208} & AGSDNet \cite{thakur2022agsdnet} & SIFSDNet \cite{thakur2022sifsdnet} & SARDDPM \cite{perera2023sar} & GUE & GUE+filter \\
			\midrule
			\multirow{3}{*}{(1)} & PSNR & 10.5148 & 8.2256 & 11.6919 & 11.8002 & 14.3954 & \textbf{15.9787}\\
			& ENL & 3.5729 & 3.9644 & 7.9241 & 6.5878 & 1.1261 & 1.2265 \\
			& SSIM & 0.0586 & 0.0559 & 0.0642 & 0.0780 & 0.1095 & \textbf{0.1271} \\
			
			\midrule
			
			\multirow{3}{*}{(2)} & PSNR & 9.7167 & 8.3171 & 11.4063 & 11.1574 & 13.8320 & \textbf{15.9811} \\
			& ENL & 3.5990 & 3.8185 & 7.7291 & 6.3417  & 0.9439 & 1.1199 \\
			& SSIM & 0.0665 &  0.0576 & 0.0667 & 0.0772 & 0.1194 & \textbf{0.1233} \\
			
			\midrule
			
			\multirow{3}{*}{(3)} & PSNR & 9.5603 & 8.5982 & 11.3086 & 11.6789 & 12.4549 & \textbf{14.9317} \\
			& ENL & 3.4256 & 3.8799 & 7.7455 & 6.5184 & 0.9461 & 1.0409 \\
			& SSIM & 0.0519 & 0.0573 & 0.0670 & 0.0773 & 0.1009 & \textbf{0.1024} \\

			\midrule
			
			\multirow{3}{*}{(4)} & PSNR & 10.4617 & 8.2644 & 11.5795 & 11.9125 & 14.5942 & \textbf{15.8834} \\
			& ENL & 3.6107 & 3.8053 & 7.8207 & 6.3731 & 1.0914 & 1.2072 \\
			& SSIM & 0.0579 & 0.0574 & 0.0664 & 0.0773 & \textbf{0.1103} & 0.1049 \\
			
			\bottomrule
		\end{tabularx}
	
\end{table*}

\begin{figure*}[!t]
		\centering
		\includegraphics[width=0.8\textwidth]{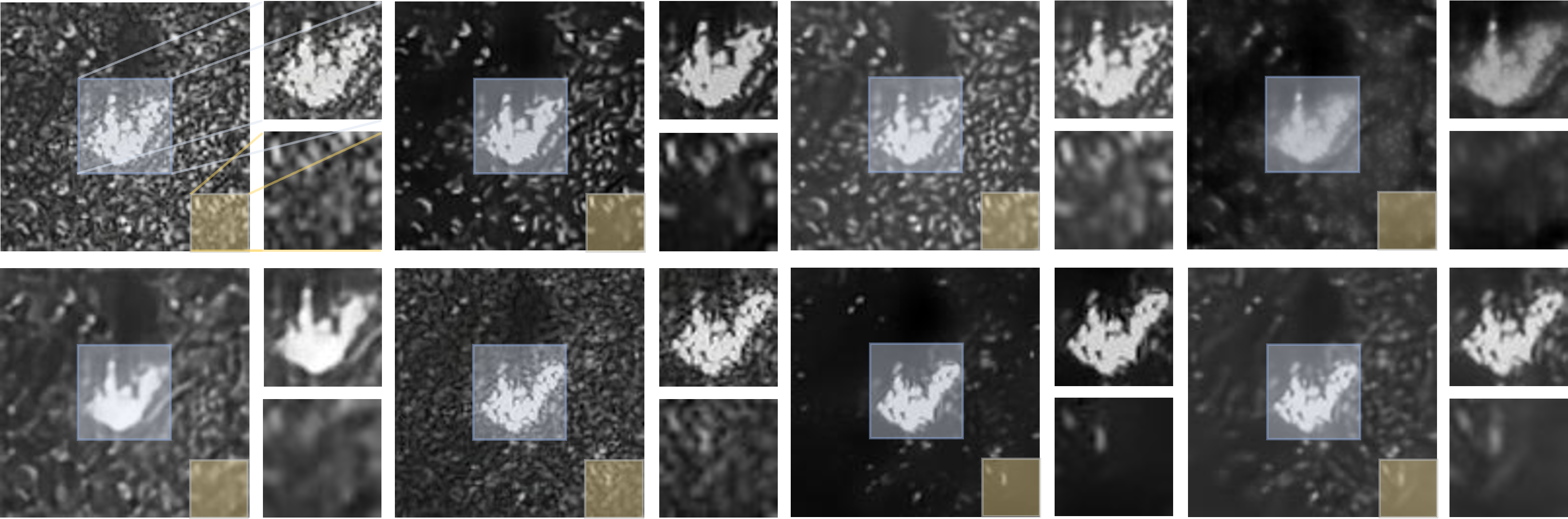}
		\caption{Detail demonstration of snythetic image despeckling, top-left to bottom-right: Origin image, ANLM, DnCNN, SAR2SAR, SARCAM, GUE, GUE+filter1, GUE+filter2.}
		\label{fig_9}
\end{figure*}

\begin{figure*}[!t]
		\centering
		\includegraphics[width=0.8\textwidth]{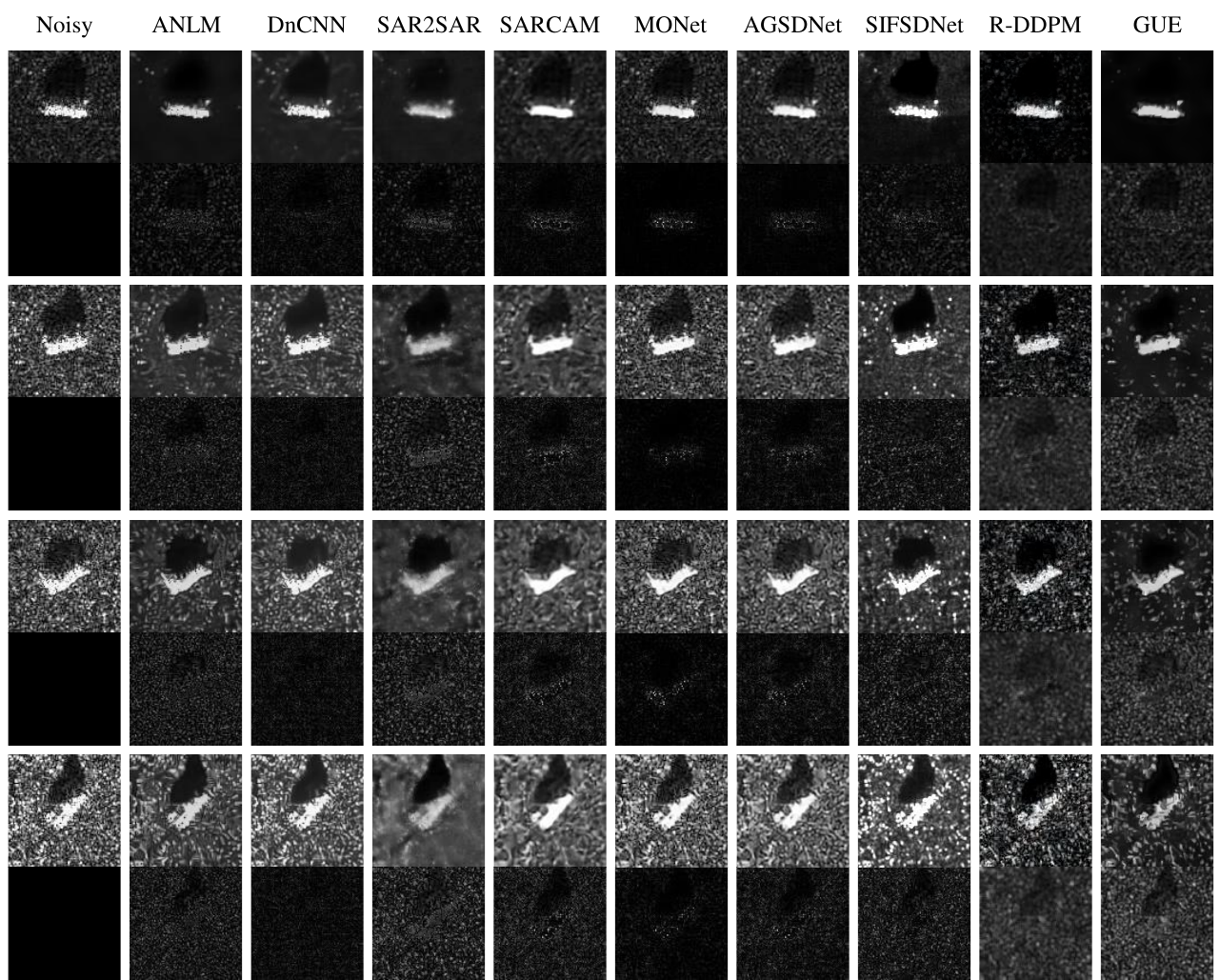}
		\caption{Real image despeckling experiment: left to right: noisy image, ANLM, DnCNN, SAR2SAR, SARCAM, MONet, AGSDNet, SIFSDNet, SARDDPM, GUE, GUE$+$filter. The original noisy image and each method contains three parts, up to down: despeckling image and residual map.}
		\label{fig_10}
	
\end{figure*}

\begin{figure*}[!t]
		\centering
		\includegraphics[width=0.8\textwidth]{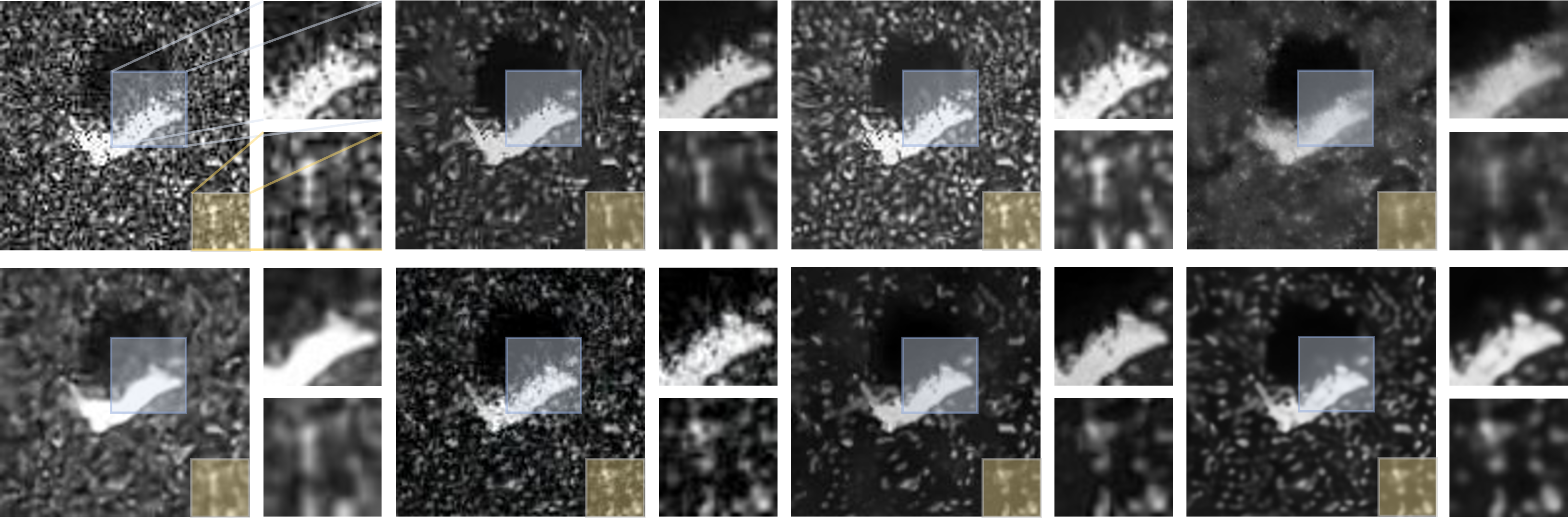}
		\caption{Detail demonstration of real image despeckling, top-left to bottom-right: Origin image, ANLM, DnCNN, SAR2SAR, SARCAM, GUE, GUE+filter1, GUE+filter2.}
		\label{fig_11}
	
\end{figure*}

\subsection{Experimental Setup}

\noindent \textbf{Dataset:} MSTAR (Moving and Stationary Target Acquisition and Recognition) dataset is a real SAR ground target dataset released by Defense Advanced Research Projects Agency (DARPA). Many domestic and foreign studies on SAR automatic target recognition and SAR image synthesis rely on this dataset. SAR in the dataset are collected using a high-resolution synthetic aperture radar with a resolution of 0.3m × 0.3m. The dataset includes sliced images of ten types of vehicles and military targets, namely 2S1, BRDM2, BTR60, D7, T62, ZIL131, ZSU234, SN132, SN9563, and SNC71.

We choose the MSTAR dataset for two reasons: (1) The MSTAR dataset contains rich semantic information, such as high and low noise samples, multiple categories of samples, and samples with various motion postures. This allows GUE to fully explore the hidden information in data and accomplish different SAR image processing tasks. (2) Due to computational limitations, we cannot train high-performance generative adversarial networks on large datasets, which would result in information loss during the reconstruction of real images.

\noindent \textbf{Preparation and Pre-trained Network:} We adopt the StyleGAN2 framework to train the generator to generate realistic SAR images and obtain the latent space information of the data. We adjust SAR image resolution to 128, select StyleGAN2 mapping network (as shown in figure \ref{fig_2}) as 8 layers, set the latent space dimension to 512, and set the number of StyleGAN blocks to 12, corresponding to $w^+$ latent space dimension (12, 512), other settings retain unchanged\footnote{ \url{https://github.com/rosinality/stylegan2-pytorch}}. In real image editing, we need to embed the real image into the latent space for editing. We first pre-train a VGG16 network on MSTAR dataset, and calculate the perceptual loss through Image2StyleGAN\footnote{\url{https://github.com/zaidbhat1234/Image2StyleGAN}}. GAN inversion has been completed. Other parameters are default parameters. 


\subsection{GUE for SAR Despeckling}\label{Section_4_2}

%

\begin{table*}[!t]
		\caption{Parameter comparison on real image(BEST INDEX VALUES ARE HIGHLIGHTED IN BOLD)}
		\centering
		\newcolumntype{C}{>{\centering\arraybackslash}X}
		\begin{tabularx}{\textwidth}{CCCCCCCC}
			\toprule
			& Metric & ANLM \cite{xiao2020asymptotic} & DnCNN \cite{zhang2017beyond} & SAR2SAR \cite{dalsasso2021sar2sar} & MONet \cite{vitale2021analysis} & InterGAN \cite{shen2020interpreting} & CFF \cite{shen2021closed} \\
			\midrule
			\multirow{3}{*}{(1)} & PSNR & 12.6183 & 10.2462 & 12.4074 & 9.6373 & 6.4239 & 10.7989 \\
			& ENL & 2.4887 & 3.0471 & 4.1884 & \textbf{6.2705} & 1.9986 & 2.3832 \\
			& SSIM & 0.0762 & 0.0748 & 0.0613 & 0.0523 & 0.0359 & 0.0410 \\
			
			\midrule
			
			\multirow{3}{*}{(2)} & PSNR & 12.7700 & 9.9121 & 13.6482 & 8.4130 & 6.4402 & 10.7740 \\
			& ENL & 2.5698 & 2.7700 & 4.2585 & \textbf{8.4932} & 2.2148 & 3.3995 \\
			& SSIM & 0.0815 & 0.0822 & 0.0656 & 0.0418 & 0.0358 & 0.0415 \\
			
			\midrule
			
			\multirow{3}{*}{(3)} & PSNR & 11.7142 & 9.1589 & 11.5931 & 9.1746 & 6.4652 & 10.8264 \\
			& ENL & 2.3639 & 2.6214 & 3.9132 & \textbf{6.6477} & 2.2356 & 3.4159 \\
			& SSIM & 0.0743 & 0.0683 & 0.0579 & 0.0403 & 0.0360 & 0.0426 \\

			\midrule
			
			\multirow{3}{*}{(4)} & PSNR & 12.8423 & 10.3721 & 12.3489 & 9.8511 & 6.5479 & 10.8695 \\
			& ENL & 2.5174 & 2.7928 & 4.2097 & \textbf{8.6403} & 2.1926 & 2.4847 \\
			& SSIM & 0.0735 & 0.0752 & 0.0638 & 0.0407 & 0.0359 & 0.0414 \\

			\bottomrule
			\label{tb2}
		\end{tabularx}
		
		\begin{tabularx}{\textwidth}{CCCCCCCC}
			\toprule
			& Metric & SARCAM \cite{9633208} & AGSDNet \cite{thakur2022agsdnet} & SIFSDNet \cite{thakur2022sifsdnet} & SARDDPM \cite{perera2023sar} & GUE & GUE+filter \\
			\midrule
			\multirow{3}{*}{(1)} & PSNR & 10.5148 & 8.2256 & 11.6919 & 11.8002 & 14.3954 & \textbf{15.9787} \\
			& ENL & 3.5729& 3.9644 & 7.9241 & 6.5878 & 1.1261 & 1.2265 \\
			& SSIM & 0.0586 & 0.0559 & 0.0642 & 0.0780 & \textbf{0.1095} & 0.1071 \\
			
			\midrule
			
			\multirow{3}{*}{(2)} & PSNR & 9.7167 & 8.3171 & 11.4063 & 11.1574 & 13.8320 & \textbf{15.9811} \\
			& ENL & 3.5990 & 3.8185 & 7.7291 & 6.3417  & 0.9439 & 1.1199 \\
			& SSIM & 0.0665 &  0.0576 & 0.0667 & 0.0772 & 0.1194 & \textbf{0.1233} \\
			
			\midrule
			
			\multirow{3}{*}{(3)} & PSNR & 9.5603 & 8.5982 & 11.3086 & 11.6789 & 12.4549 & \textbf{14.9317} \\
			& ENL & 3.4256 & 3.8799 & 7.7455 & 6.5184 & 0.9461 & 1.0409 \\
			& SSIM & 0.0519 & 0.0573 & 0.0670 & 0.0773 & 0.1009 & \textbf{0.1024} \\

			\midrule
			
			\multirow{3}{*}{(4)} & PSNR & 10.4617 & 8.2644 & 11.5795 & 11.9125 & 14.5942 & \textbf{15.8834} \\
			& ENL & 3.6107 & 3.8053 & 7.8207 & 6.3731 & 1.0914 & 1.2072 \\
			& SSIM & 0.0579 & 0.0574 & 0.0664 & 0.0773 & \textbf{0.1103} & 0.1049 \\

			\bottomrule
			\label{tb2}
		\end{tabularx}
\end{table*}

%

\noindent \textbf{Evaluation Metric: }We employed structural similarity (SSIM), peak signal-to-noise ratio (PSNR), equivalent number of looks (ENL) as evaluation metrics for the MSTAR dataset. To obtain the ground truth noise-free images, we annotated a portion of data and frame the target region. According to \cite{ma2022no}, PSNR is a commonly used measurement method for image reconstruction, defined by the mean square error, and serves as an image similarity index, and SSIM calculates the similarity between two images, providing a measure of the noise level in the reconstructed denoised image \cite{ma2022no}.

\noindent \textbf{Comparison Models: }MSTAR dataset can represent a specific scenario's despeckling requirements, demanding algorithms to have better adaptability for removing mixed noise. We compared GUE with other despeckling methods on this dataset. Considering the challenging nature of mixed noise within the dataset, we selectively chose open-source methods for testing, aiming to compare their despeckling capability and ability to retain image details. The methods employed in the comparative experiments include ANLM \cite{xiao2020asymptotic}, DnCNN \cite{zhang2017beyond}, SAR2SAR \cite{dalsasso2021sar2sar}, SAR-CAM \cite{9633208}, MONet \cite{vitale2021analysis}, Inter-GAN \cite{shen2020interpreting} Closed-Form Factorization (CF-Factor) \cite{shen2021closed}, AGSDNet \cite{thakur2022agsdnet}, SIFSDNet \cite{thakur2022sifsdnet} and SARDDPM \cite{perera2023sar}. In SAR-CAM, we utilized the parameter settings recommended in the paper for training. In DnCNN, the noise level used during training was set to $\sigma = 75$. Following the hyperparameter index in the original article, we set the search range in each direction to 8 and the image block radius to 2 when performing ANLM, given that the image resolution of our dataset is 128. Similarly, for SAR2SAR method, the image was divided into 64x64 patches with a stride of 8. For InterGAN, we trained a support vector machine to obtain the noise hyperplane and operated latent code along the normal vector of the hyperplane to obtain despeckling results. Other methods were tested using their pre-trained weights.

\noindent \textbf{Synthetic Image Experiments: }Figure \ref{fig_7} shows despeckling semantic in GANs' latent space. We configured GUE to search for 512 directions and generated a random normal distribution $z$ vector with a dimension of $[1, 512]$. Due to the characteristics of GUE's unsupervised learning, it identifies the first 200 directions representing the most significant change. After the artificial selection, some directions are identified as potential noise semantic directions. And we can always find a corresponding noise semantic direction for different SAR image samples for SAR despeckling.

Our objective was to identify the optimal despeckling method that can effectively remove background speckle noise while preserving the details of target area. To thoroughly assess each algorithm's detail preservation and despeckling abilities, we selected multiple images with varying noise levels for testing. We manually annotated and framed the target areas, extracting noise-free images for evaluation metrix calculations.

Figure \ref{fig_8} presents the comparison results of each method. For low-level speckle noise, ANLM performs well in despeckling and detail preservation. DnCNN exhibits weak despeckling ability, while SAR2SAR and SARCAM suffer significant loss of target details. InterGAN and CFF give poor results and inevitably causes deformation during the despeckling process. GUE effectively reduces background noise without compromising details when processing low-noise images, and it can eliminate nearly all background noise from weak-speckle images after filtering. In the despeckling of high-noise images, the methods used in the comparative experiments have minimal impact. In contrast, GUE still manages to reduce the background speckle noise of SAR images to a certain extent, showcasing superior performance after filtering. By leveraging the semantic features of GAN latent space for despeckling, GUE's performance remains unaffected by the intensity of image background speckle noise. Figure \ref{fig_9} presents an enlarged view depicting the details of various methods, demonstrating its superior performance compared to other methods. Table \ref{tb1} presents the quantitative evaluation of GUE in terms of despeckling. We selected four sets of images and conducted experiments on 50 randomly chosen images from each set. The results indicate that GUE outperforms other existing methods in most metric except for the ENL metric.

\noindent \textbf{Real Image Experiments: }To facilitate the application of GUE for despeckling on authentic images, it is necessary to initially map the authentic images to the latent space of GAN. Upon comparing the reconstruction outcomes in $z$ space, $w$ space, and $w^{+}$ space in figure \ref{fig_2}, we observed that the reconstructed images in $z$ space and $w$ space are often clear but prone to deformation. Consequently, we opted to conduct image reconstruction in $w^{+}$ space. We inverted MSTAR dataset to GANs' latent space and employed GUE to identify latent despeckling semantic directions for image editing. We configured GUE to search for 512 directions but abstained from utilizing a random normal distribution as the initial noise input for GUE based on $w^{+}$ space. After GUE converges, we screen for despeckling semantic orientations and use orientation-aware SAR image despeckling in the following experiments.

Similar to the synthetic image experiment, we randomly selected several samples to create a dataset for labeling. We utilized PSNR, SSIM, and ENL as the measurement indicators. Figure \ref{fig_10} compares the results obtained by each method for processing authentic SAR images. Kernel density estimation reduces the influence of the reconstructed data distribution on the semantic direction to a certain extent. Denoising the SAR image along the discovered semantic direction hardly causes deformation within a specific range while preserving the details of the target area as much as possible. GUE performs comparably to state-of-the-art methods in low-noise image despeckling and achieves excellent results in heavy-speckle image despeckling. Table \ref{tb2} presents the quantitative comparison results of each method. Considering the brightness change caused by image reconstruction and the slight reduction in the parameter index of GUE, it is anticipated that future advancements in GAN inversion technology will further enhance the performance of this method. By incorporating a filtering operation after GUE, optimal performance can suppress background noise and preserve target details.

\subsection{GUE for SAR Background Segmentation}\label{Section_4_3}

\noindent \textbf{Evaluation Metric: }We adopted two quantitative evaluation metrics, Dice Score ($S_{dice}$) \cite{wang2020solo} and Mean Pixel Accuracy (MPA), based on previous research. The definition of Dice Score is as follows:

\begin{equation}
	S_{\text {dise }}(p, q)=\frac{2 \sum_{x, y}\left(p_{x, y} \cdot q_{x, y}\right)}{\sum_{x, y} p_{x, y}^2+\sum_{x, y} q_{x, y}^2},
\end{equation} where $p_{x,y}$ and $q_{x,y}$ are the pixel positions in segmentation result and the Ground-truth, respectively. Subsequently, we define MPA as:

\begin{equation}
	M P A=\frac{1}{2} \sum_{i=1}^n \frac{p_{i j}}{\sum_{j=1}^n p_{i j}},
\end{equation} where $p_{ij}$ represents the number of pixels of target $j$ that are identified as background $i$.

For synthetic image experiments, we manually annotated 200 samples for experimentation. For real image editing, we utilized the SARBake dataset \cite{malmgren2015convolutional} as the ground-truth.

\begin{figure}[!t]
	\centering
	\includegraphics[width=\linewidth]{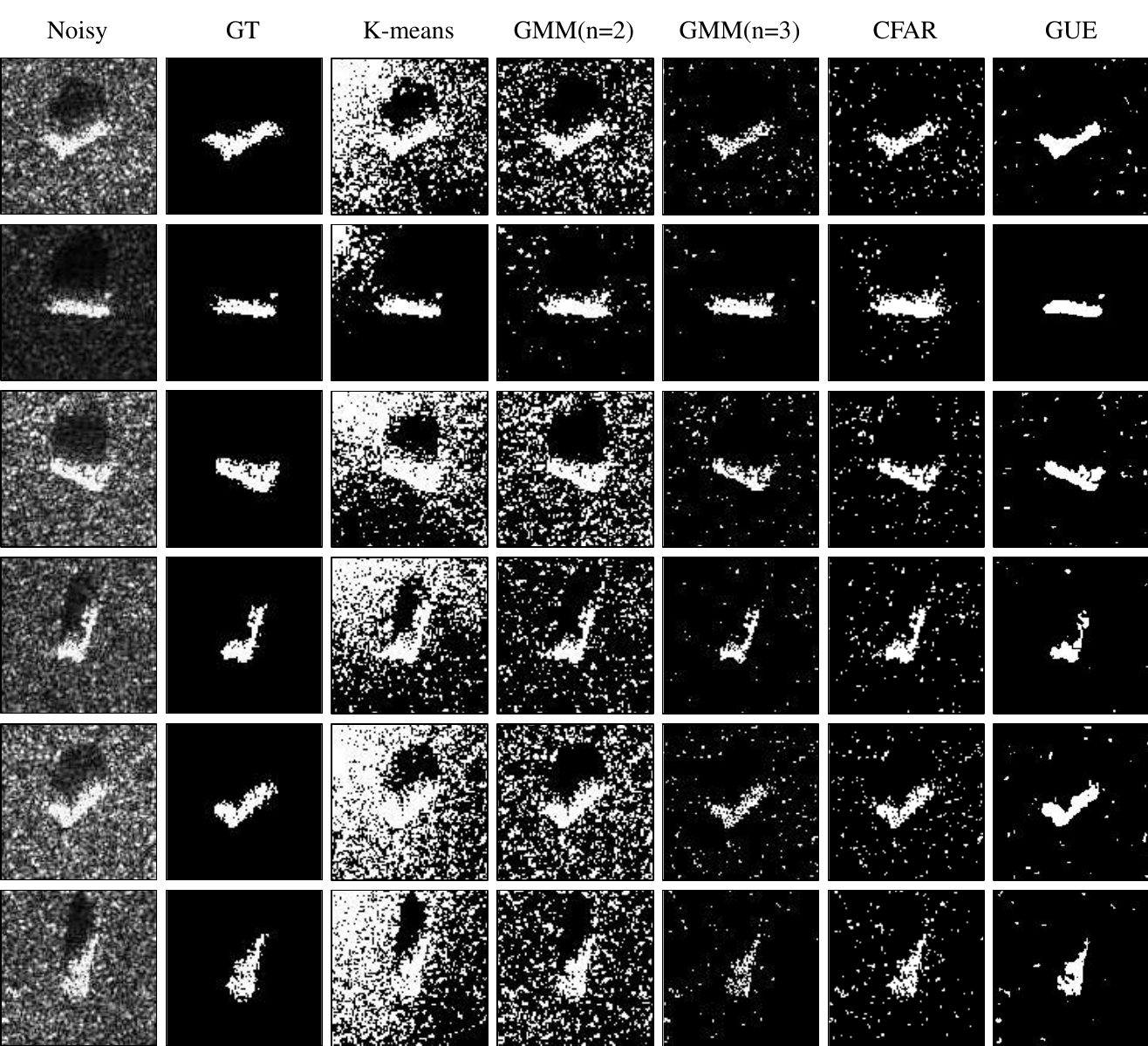}
	\caption{Real image segmentation experiment: left to right: noisy image, ground truth, K-means, GMM(n=2), GMM(n=3), CFAR, GUE.}
	\label{fig_13}
\end{figure}

\begin{table}[!t]
	\caption{{Synthetic image segmentation experiment on MSTAR datasets(BEST INDEX VALUES ARE HIGHLIGHTED IN BOLD).}}\label{tb3}
	\newcolumntype{C}{>{\centering\arraybackslash}X}
	\begin{tabularx}{\linewidth}{p{2.6cm}CC}
		\toprule
		\textbf{Method} & \textbf{$S_{dice}$$\uparrow$} & \textbf{$MPA$$\uparrow$} \\
		\midrule
		K-means \cite{zhang2008spectral} & 0.3103 & 0.528 \\
		GMM(n=2) \cite{belloni2017sar} & 0.4176 & 0.435 \\
		GMM(n=3) \cite{belloni2017sar} & 0.6291 & 0.667 \\
		CFAR \cite{tao2016segmentation} & 0.6741 & 0.866 \\
		GUE & \textbf{0.7725} & \textbf{0.907} \\
		\bottomrule
	\end{tabularx}
\end{table}

\noindent \textbf{Comparison Models: }We evaluated several open-source segmentation algorithms on MSTAR dataset, including clustering methods \cite{zhang2008spectral}, CFAR \cite{tao2016segmentation}, GMM \cite{belloni2017sar}, ASC \cite{gerry1999parametric, feng2021target}, and ACM Net \cite{feng2021target}. To ensure fairness in comparison, we only selected unsupervised segmentation algorithms here, without including supervised algorithms that require annotation. In our specific implementation, we utilized K-means clustering for SAR background segmentation with the number of clusters set to 2. For Gaussian Mixture Model (GMM), we set the maximum iteration to 1000 and conducted experiments with 2 and 3 clusters. In CFAR, we defined the target size as 30x30 and set the hyperparameter Pfa to 0.01. For comparison with Mask R-CNN \cite{he2017mask} and ACM Net, we directly referred to the data in \cite{feng2021target} and contrasted it with the experimental results of GUE on real images.

\begin{figure}[!t]
	\centering
	\includegraphics[width=\linewidth]{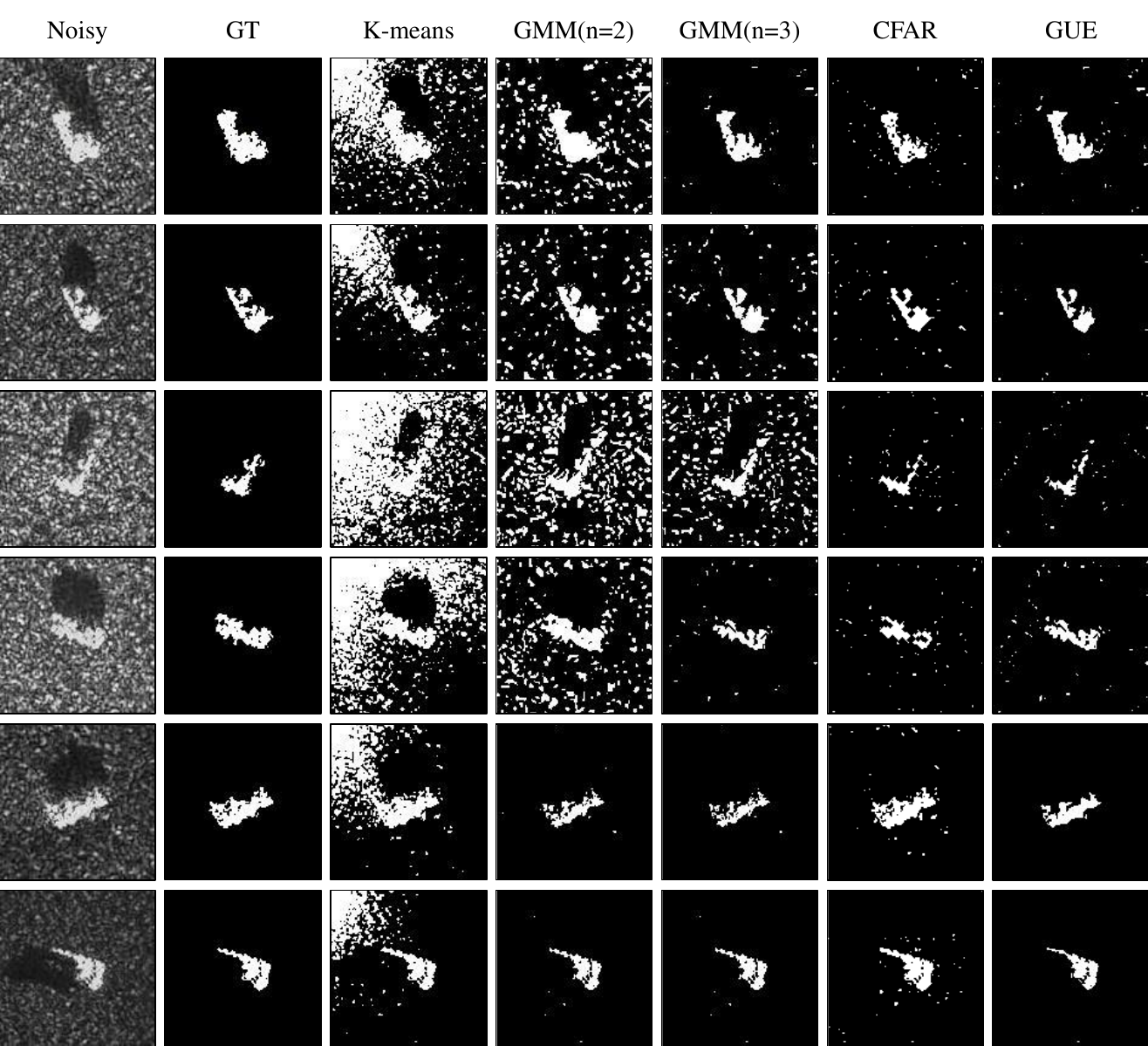}
	\caption{Synthetic image segmentation experiment: left to right: noisy image, ground truth, K-means, GMM(n=2), GMM(n=3), CFAR, GUE.}
	\label{fig_12}
\end{figure}

\noindent \textbf{Experiments on MSTAR Datasets: }The advantage of GUE lies in its ability to segment samples with extremely high background noise. Since this method delineates between background and target regions by seeking noise semantics in latent space, GUE is unaffected by the choice of hyperparameters and can discern target and background regions in complex background information. Figure \ref{fig_12} illustrates the visual results of different SAR segmentation methods on synthetic images, where images with varying degrees of background noise were tested. It can be observed that while K-means can obtain target regions, it also retains some noise in background. In the case of GMM, experiments were conducted with clustering numbers set to 2 and 3. While GMM fails to effectively segment images with high background noise with a clustering number of 2, there is a significant improvement in performance with a clustering number of 3, albeit requiring subsequent filtering. CFAR demonstrates good performance in SAR image segmentation with appropriate hyperparameter, achieving segmentation of images with varying degrees of noise. GUE suppresses background noise by moving latent code along the low-noise semantic direction in GANs' latent space, followed by retaining target regions through thresholding, resulting in satisfactory segmentation results. Table \ref{tb3} provides quantitative analysis results of different methods on synthetic images, indicating that both GUE and CFAR outperform other methods. 

\begin{table}[!t] 
	\caption{{Real image segmentation experiment on MSTAR datasets(BEST INDEX VALUES ARE HIGHLIGHTED IN BOLD).}}\label{tab4}
	\newcolumntype{C}{>{\centering\arraybackslash}X}
	\begin{tabularx}{\linewidth}{p{2.6cm}CC}
		\toprule
		\textbf{Method} & \textbf{$S_{dice}$$\uparrow$} & \textbf{$MPA$$\uparrow$} \\
		\midrule
		K-means \cite{zhang2008spectral} & 0.4121 & 0.386 \\
		GMM(n=2) \cite{belloni2017sar} & 0.4262 & 0.481 \\
		GMM(n=3) \cite{belloni2017sar} & 0.6514 & 0.746 \\
		CFAR \cite{tao2016segmentation} & 0.5780 & 0.674 \\
		Mask R-CNN \cite{he2017mask} & 0.7433 & 0.892 \\
		ACM Net \cite{feng2021target} & 0.7521 & \textbf{0.894} \\
		GUE & \textbf{0.7544} & 0.884 \\
		\bottomrule
	\end{tabularx}
\end{table}

In real image experiments, effective segmentation of target and background regions can be achieved for different levels of noise. Compared to synthetic image experiments, GUE also achieves similar performance in real image experiments. However, since GUE requires the restoration of real images to the GAN's $w^+$ space, this can lead to blurring and loss of some target details in the images. Figure \ref{fig_13} presents a comparison of segmentation results for real images, while Table \ref{tb4} provides corresponding quantitative analysis results.

\begin{figure*}[!t]
	\centering
	\includegraphics[width=\linewidth]{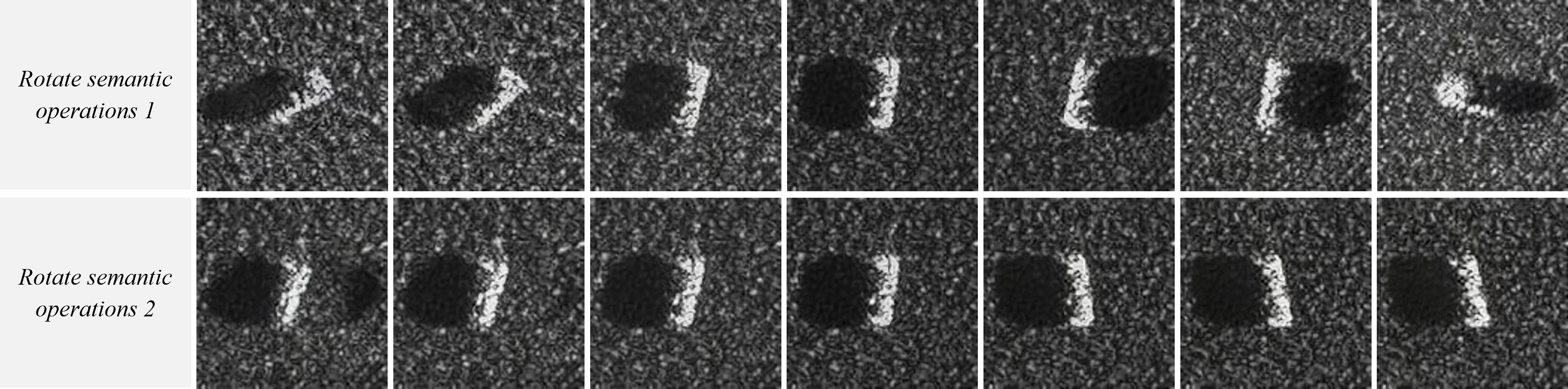}
	\caption{Rotation semantic experiment: GUE finds the rotation of the target along different axes and can simultaneously operate the transformation of the interference spot.}
	\label{fig_14}	
\end{figure*}

\subsection{GUE for SAR Rotation and Data Enhancement}\label{Section_4_4}

\begin{table}[!t]
	\caption{Rotational Semantic Experiment: Rotational semantic direction from different classes ensures their categories remain unchanged.}
	\label{tb5}
	\centering
	\setlength{\tabcolsep}{4pt}
	\begin{tabularx}{\linewidth}{YYYYYY} 
		\toprule
		Class & Direction & Conf. & Class & Direction & Conf. \\
		\midrule
		\multirow{3}{*}{2S1(1)} & 65 & 0.96024 & \multirow{3}{*}{2S1(2)} & 65 & 0.99488 \\
		& 200 & 0.97999 & & 200 & 0.98446 \\
		& 393 & 0.98987 & & 393 & 0.99701 \\
		\midrule
		\multirow{3}{*}{BRDM2(1)} & 23 & 0.99620 & \multirow{3}{*}{BRDM2(2)} & 23 & 0.99995 \\
		& 148 & 0.99816 & & 148 & 0.96177 \\
		& 460 & 0.99995 & & 460 & 0.99860 \\
		\midrule
		\multirow{3}{*}{T62(1)} & 16 & 0.34855 & \multirow{3}{*}{T62(2)} & 16 & 0.97624 \\
		& 133 & 0.88160 & & 133 & 0.85891 \\
		& 432 & 0.85005 & & 432 & 0.86683 \\
		\midrule
		\multirow{2}{*}{SNC71(1)} & 94 & 0.99076 & \multirow{2}{*}{SNC71(2)} & 94 & 0.94459 \\
		& 417 & 0.95702 & & 417 & 0.99776 \\ 
		\midrule
		\multirow{2}{*}{ZSU234(1)} & 19 & 0.92353 & \multirow{2}{*}{ZSU234(2)} & 19 & 0.72468 \\
		& 258 & 0.55991 & & 258 & 0.95857 \\   
		\bottomrule
	\end{tabularx}
\end{table}

\noindent \textbf{Experiments on MSTAR Datasets: }GUE has proven effective in identifying various semantic directions in SAR images, including both noise-related and human-interpretable directions, such as rotation and category transformation. When a semantic direction results remain unchanged in category after manipulation, it is defined as a rotation semantic direction. To ensure the validity of the rotational semantic direction, we imposed specific constraints on the ENL (Equivalent Number of Looks) and mean value of the manipulated image. These constraints are necessary because some noise semantic orientations may be coupled with the rotational semantic direction, and we want to ensure the distinction.


Traditional SAR image editing techniques often focused on altering the target area while leaving the interference spots near the target unchanged. These approaches led to the loss of critical image information and hindered the effective processing of target and interference spot areas. However, by leveraging the rotation semantic direction identified by GUE to operate the latent code, we can simultaneously modify the shape of both the target and interference spot areas. This innovative approach allows for collaborative target and interference spots information processing, resulting in more accurate and informative image manipulations.


Figure \ref{fig_14} visually illustrates different rotation semantic directions on the same sample. Table \ref{tb5} provides additional examples of rotation semantic directions. We observed that samples from various categories exhibit distinct sensitivities to semantic directions. Semantic directions identified in specific categories (e.g., 2S1, BRDM\_2) can effectively implement rotation for most samples. However, samples from some categories (such as BTR\_60) can only achieve effective rotation editing along a fixed direction. Due to the interweaving of GAN latent space features, some samples may remain semantic coupling even when simplified from high-dimensional features to data manifolds. Nevertheless, GUE consistently discovers adequate rotation semantics for SAR image editing. Importantly, after passing through a pre-trained ResNet, the top-1 confidence categories of these directions remained unchanged, which confirms the accuracy of the rotation semantic directions discovered by GUE, including human-interpretable and visually incomprehensible orientations. Finding semantic directions that rotate along different axes enhances the perceptibility and intuitiveness of SAR target rotation imaging for human observers. This demonstrates the potential of GUE in advancing SAR image processing and interpretation.

\begin{figure*}[!t]
	\centering
	\includegraphics[width=18cm]{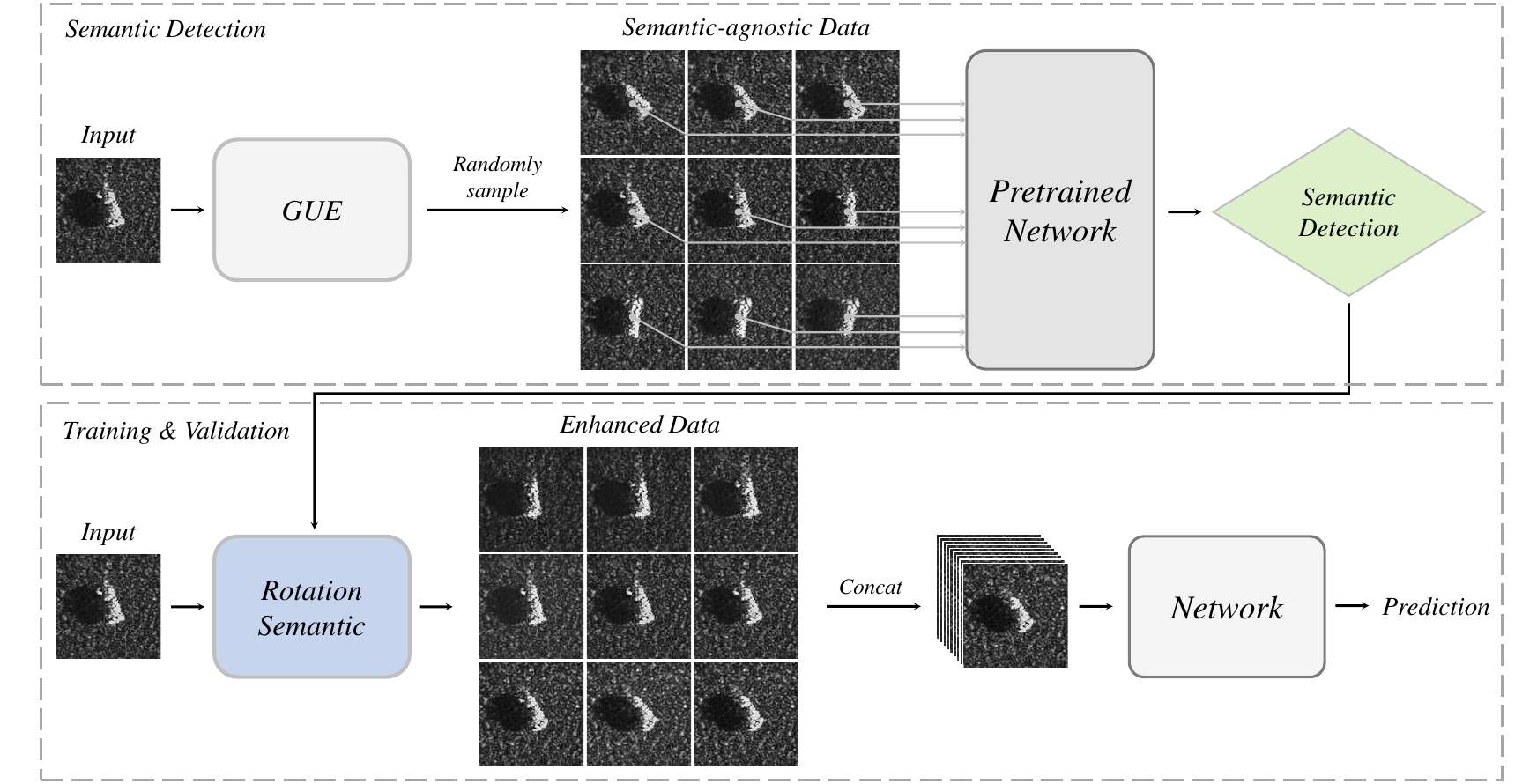}
	\caption{Semantic enhancement involves leveraging the semantic information extracted by GUE to the fullest extent by establishing multi-channel inputs. This is achieved by modifying the input layers of various baseline networks and subsequently conducting training and testing.}
	\label{fig_15}
\end{figure*}

\subsection{GUE for Guided SAR Target Recognition}\label{Section_4_5}

\noindent \textbf{Implementation Specifics}: To fully leverage correlated semantic information to guide recognition task, we modified network architecture to accommodate multi-channel network inputs. Specifically, in our experiments, we altered the first layer of the network to have 9-channel inputs while keeping the rest of network structure unchanged. In continuous semantic enhancement, for each training sample, we randomly generated 8 augmented samples using GUE's rotation semantic direction and combined them with the original sample to form a 9-channel input for training. In comparative experiments, we utilize the network's average confidence obtained from the baseline. Figure \ref{fig_15} illustrates the structures of both traditional network and GUE-guided multi-channel data enhancement network. By utilizing GUE to explore rotation semantic directions, we integrated continuous semantic multi-channel network inputs to guide SAR recognition. We conducted tests on various baseline networks.

\begin{table}[!t]
		\caption{{Comparison results between GUE-guided SAR ATR and baselines (BEST INDEX VALUES ARE HIGHLIGHTED IN BOLD).}}\label{tb6}
		\newcolumntype{C}{>{\centering\arraybackslash}X}
		\begin{tabularx}{\linewidth}{p{3cm}CC}
			\toprule
			Backbone & Baseline & Ours \\
			\midrule
			AlexNet & 0.952 & \textbf{0.968} \\
			VGG16 & 0.973 & \textbf{0.991} \\
			ResNet18 & 0.984 & \textbf{0.990} \\
			ShuffleNetv2 & 0.961 & \textbf{0.984} \\
			MobileNetv2 & 0.980 & \textbf{0.982} \\
			AConvNet & 0.979 & \textbf{0.983} \\
			AM-CNN & \textbf{0.977} & 0.975 \\
			BagNet17 & 0.970 & \textbf{0.982} \\
			\bottomrule
		\end{tabularx}
\end{table}

\noindent \textbf{Experiments on MSTAR Datasets: }We proposed GUE-guided SAR ATR by introducing multi-channel inputs to the backbone network. This modification allowed for a more comprehensive representation of SAR images, capturing diverse semantic information crucial for accurate target recognition. The inclusion of multi-channel inputs enabled the network to consider a wider range of features, including variations in orientation, scale, and context, which are essential for distinguishing between different classes in complex SAR scenes. Table \ref{tb6} presents a comprehensive comparison between various baseline networks and the enhanced network with GUE guidance. Notably, the introduction of continuous rotation semantics further augmented the network's feature extraction capabilities. By integrating semantic information related to rotation, the network became adept at capturing subtle variations in target appearance caused by changes in orientation, leading to improved recognition performance. Moreover, the incorporation of multi-channel inputs facilitated a more robust and adaptable learning process within the convolutional network architecture. The network could effectively leverage the additional information provided by each channel to refine its internal representations and enhance its discriminative power. This enabled the network to achieve superior performance across different SAR datasets, demonstrating the effectiveness of GUE-guided approach in enhancing SAR ATR capabilities.

\subsection{Ablation Study}\label{Section_4_6}

In this section, we design ablation experiments to verify the effectiveness of designs in GUE. Specifically, we prove the effectiveness of the $L$ regression term in the optimization model and ascertain the necessity of introducing kernel density estimation for model training. Table \ref{tb4} gives the specific results of the ablation experiments.

\begin{figure}[!t]
	\centering
	\includegraphics[width=0.9\linewidth]{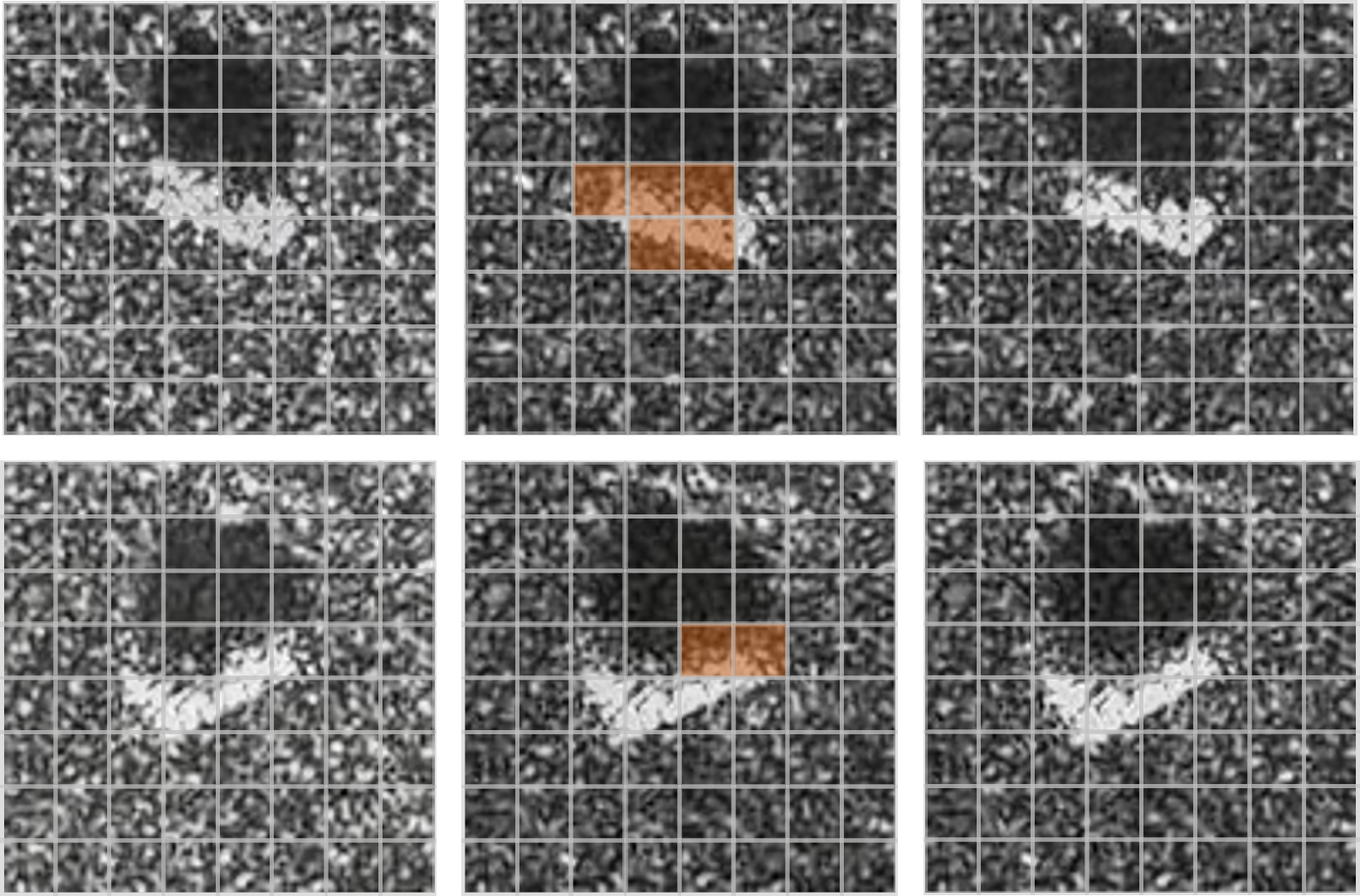}
	\caption{Ablation study for $L$ loss term: column: Left to right: Origin image, GUE without $L$ loss, GUE. The red blocks mark the distorted regions.}
	\label{fig_16}
\end{figure}

\noindent \textbf{$L$ regression term:} We found that the optimization speed of the model was languid during the synthetic image experiment, and the directions were not completely separated after long-term training. Therefore, we add the displacement distance term and observe that the training speed of the model is significantly improved. However, the loss corresponding to the $L$ regression term is almost unchanged, and only the $H$ regression term controls the overall loss's decline. It is currently known that if the operation vector in the latent space of GAN changes continuously, the output image of the generator will also change continuously, so we infer that the introduction of the displacement distance item makes the direction change continuously, which makes it possible to find meaning in the latent space of GAN. Figure \ref{fig_16} shows despeckling effect before and after removing the $L$ regression item. The $L$ regression term ensures the stability of model training. Moreover, including the $L$ regression term facilitates the identification of a semantic direction that is easier to comprehend.

\noindent \textbf{Kernel density estimation:} Refer to the estimation methods in \cite{zandieh2023kdeformer}. We tried to directly apply the semantic operation operator in the synthetic image experiment to the actual image after inversion and found that we could only achieve a low level of despeckling. Randomly initializing the latent code through a normal distribution will produce a fuzzy generator output. To make the semantic operator suitable for the data distribution of the $w^{+}$ space, we introduce kernel density estimation to generate $w$ random codes. Kernel density estimation effectively solves the problems of blurring and brightness changes caused by $w^{+}$ space despeckling.

\begin{table*}[]
	\centering
	\caption{Ablation studies. The despeckling indicators of different module configurations are evaluated on MSTAR. The symbol "\cmark" means used in the model, and the symbol "\xmark" means not used. The best results are marked in \textbf{bold}.\label{tb4}}
	\label{tab:ablation_study}
	\begin{tabularx}{18cm}{YYYYYYYYYYYY}
		\toprule[1pt]
		\multirow{2}{*}{} & \multirow{2}{*}{KDE} & \multicolumn{3}{c}{Selection of $F$} & \multirow{2}{*}{$L$ loss} & \multicolumn{3}{c}{Operating Distance} & \multirow{2}{*}{PSNR} & \multirow{2}{*}{SSIM} & \multirow{2}{*}{IoU(avg)} \\ \cmidrule{3-5} \cmidrule{7-9}
		& & Linear & Network & Orthogonal & & $5\alpha$ & $10\alpha$ & $30\alpha$ & & & \\ 
		\midrule
		(a) & \cmark & \xmark & \xmark & \cmark & \cmark & \xmark & \xmark & \cmark & 9.3445 & 0.0414 & 0.1672 \\
		(b) & \cmark & \xmark & \xmark & \cmark & \cmark & \cmark & \xmark & \xmark & 10.8342 & 0.0480 & 0.2016 \\
		(c) & \xmark & \xmark & \xmark & \cmark & \cmark & \xmark & \cmark & \xmark & 6.1932 & 0.0296 & 0.1127 \\
		(d) & \cmark & \xmark & \xmark & \cmark & \xmark & \xmark & \cmark & \xmark & 9.1842 & 0.0402 & 0.1579 \\
		(e) & \cmark & \cmark & \xmark & \xmark & \cmark & \xmark & \cmark & \xmark & 7.4731 & 0.0328 & 0.1221 \\
		(f) & \cmark & \xmark & \cmark & \xmark & \xmark & \xmark & \cmark & \xmark & 7.8644 & 0.0337 & 0.1295 \\
		(g) & \cmark & \xmark & \xmark & \cmark & \cmark & \xmark & \cmark & \xmark & \textbf{11.2253} & \textbf{0.0699} & \textbf{0.2750} \\ 
		\bottomrule[1pt]
	\end{tabularx}
\end{table*}

\section{Discussion}\label{Section_5}

\subsection{Core Idea}

GUE achieves SAR image editing by searching for meaningful semantic directions in the latent space of GAN, representing a completely unsupervised image processing method. The core idea of GUE is to analyze the distribution of data and reduce high-dimensional features to a low-dimensional space for editing operations, requiring the data distribution of samples to contain corresponding semantics. GUE demonstrates enormous potential in the field of SAR image editing, enabling flexible manipulation of various features in SAR images. This allows researchers to gain a more intuitive understanding of SAR imaging mechanisms and guide SAR image despeckling and recognition tasks.

\subsection{Limitation}\label{Section_5_1}

\noindent \textbf{Selection of data:} While GUE has demonstrated exceptional performance in despeckling heavily speckle SAR images, it still fails to produce satisfactory results in specific scenarios. To ensure that the latent space of GAN encompasses noise semantics, it is essential that the dataset contains images with varying levels of noise. If the noise levels in the dataset's samples are mainly uniform, the GAN will fail to capture the nuances of noise semantics. In such cases, it becomes necessary to revise the dataset by augmenting it with denoised and noisy data and then retraining the GAN. This process transforms the GAN model into a self-supervised/supervised learning model, enabling it to explore latent semantic directions through GUE.

\noindent \textbf{Limitations of GAN inversion:} Simultaneously, it is worth noting that the current GAN inversion algorithms cannot fully restore all image details in complex scenes, especially for SAR images. When utilizing the $w^+$ method for image reconstruction, the reconstruct image tends to exhibit a certain degree of blurring that can be challenging to eliminate. Additionally, the presence of artifacts can introduce interference in the final output. However, these issues are expected to be addressed and resolved as GAN inversion technology advances.

\subsection{Future Work}\label{Section_5_2}

Given the aforementioned limitations, potential future research include the efficient search for paths within GANs' latent space. Since the multilayer perceptron module of StyleGAN cannot efficiently learn the data distribution, semantic paths in the latent spaces $w$ and $w^+$ may not be linear. As a result, linear paths can only achieve interpretable semantic editing for short distances. To achieve long-distance semantic dependencies, one could explore curved paths within the latent space or replace MLP module within GAN that better learn the data distribution and achieve pre-decoupling. Additionally, improving GAN inversion methods to allow for more effective real image editing is another promising future direction.


\section{Conclusion}\label{Section_6}

This paper introduces an innovative multi-task SAR image processing framework, termed GAN-based Unsupervised Editing (GUE), which is the first attempt to manipulate semantic latent code in GANs' latent space for SAR image processing. We propose a meticulously designed decoupling network that enables both the decoupling of GAN latent space and extraction of interpretable semantic directions, which are then applied to various SAR image processing tasks including despeckling, segmentation, and guided SAR target recognition. In SAR despeckling, GUE outperforms existing supervised learning methods. Additionally, GUE allows for the synchronized rotation of SAR targets and interference speckles, addressing the issue of previous methods neglecting reflective properties. We also demonstrate that GUE can guide SAR ATR tasks effectively. Our main contribution is the development of a prototype for an unsupervised, comprehensive model that achieves multiple tasks with a single training process and delivers state-of-the-art performance. Currently, GUE allows for short-range semantic manipulations, and the embedding of real images can somewhat impact the model's performance. Potential future work includes exploring long-distance semantic manipulation in GANs' latent space and improving SAR inversion methods to further enhance image editing and processing capabilities.

\bibliographystyle{IEEEtran}
\bibliography{reference}

\end{document}